%% file: paper.tex
\documentclass{article}

\usepackage[preprint]{neurips_2026}

\input{headers/header-lqa-head}

\input{headers/header}

\input{headers/header-lqa-tail}

\DeclareMathOperator{\STEP}{STEP}

\setlist[itemize]{noitemsep, topsep=0pt}

\usetikzlibrary{shapes.geometric,circuits.logic.US,calc,positioning}
\usepackage{pifont}
\usepackage{placeins}
\usepackage{algorithm}
\usepackage{algorithmic}
\Crefname{figure}{Fig.}{Figs.}
\def\gE{\mathcal{E}}
\usepackage{multirow}

\usepackage{threeparttable}
\usepackage{enumitem}
\usepackage{booktabs}

\title{Learning Compact Boolean Networks}

\author{%
  Shengpu Wang\textsuperscript{1} \quad Yuhao Mao\textsuperscript{2} \quad Yani Zhang\textsuperscript{1} \quad Martin Vechev\textsuperscript{2} \\
  {\small \textsuperscript{1}Department of Information Technology and Electrical Engineering \quad \textsuperscript{2}Department of Computer Science} \\
  ETH Zurich \\
  {\footnotesize\texttt{wangshen@ethz.ch, yanizhang@mins.ee.ethz.ch, \{yuhao.mao,martin.vechev\}@inf.ethz.ch}}
}

\begin{document}

\maketitle

\begin{abstract}
\input{abstract.tex}
\end{abstract}

\input{sections/introduction.tex}

\input{sections/related_work.tex}

\input{sections/background.tex}

\input{sections/method.tex}

\input{sections/experiment.tex}

\input{sections/discussion.tex}
\input{sections/conclusion.tex}

\FloatBarrier
\newpage

\bibliographystyle{plainnat}
\bibliography{references}

\appendix

\input{sections/appendix.tex}

\end{document}

%% file: headers/header-lqa-head.tex
\usepackage[utf8]{inputenc} %

\usepackage{graphicx}

\usepackage[T1]{fontenc}

\input{headers/lqa/overfull}

\input{headers/lqa/comments}

\input{headers/lqa/appendix}

\input{headers/lqa/abbreviations}

\input{headers/lqa/colors}

\input{headers/lqa/listing}

\input{headers/lqa/tikz}

%% file: headers/lqa/overfull.tex
\IfFileExists{headers/config/showoverfull.config}{
	\overfullrule=1cm
}{
}

%% file: headers/lqa/comments.tex
\usepackage{marginnote}

\usepackage[backgroundcolor=none,linecolor=red,textsize=footnotesize]{todonotes}

%% file: headers/lqa/appendix.tex
\usepackage{etoolbox}

\newbool{includeappendix}
\setbool{includeappendix}{true} %
\IfFileExists{headers/config/noappendix.config}{
	\setbool{includeappendix}{false}
}{}

\newif\ifincludeappendixx
\ifbool{includeappendix}{
	\includeappendixxtrue
}{
	\includeappendixxfalse
}

\usepackage{xr} %
\usepackage{filecontents}

\ifbool{includeappendix}{}{
	\input{appendix-labels-loader}

	\externaldocument{appendix-labels}
}

%% file: headers/lqa/abbreviations.tex
\newcommand{\eg}{e.g., }
\newcommand{\ie}{i.e., }

%% file: headers/lqa/colors.tex
\definecolor{my-full-blue}{HTML}{1F77B4}

\definecolor{my-full-orange}{HTML}{FF7F0E}

\definecolor{my-full-green}{HTML}{2CA02C}

\definecolor{my-full-red}{HTML}{d62728}

\definecolor{my-full-purple}{HTML}{9467bd}

\definecolor{my-full-brown}{HTML}{8c564b}

\definecolor{my-full-pink}{HTML}{e377c2}

\definecolor{my-full-gray}{HTML}{7f7f7f}

\definecolor{my-full-olive}{HTML}{bcbd22}

\definecolor{my-full-cyan}{HTML}{17becf}

\definecolor{muted-green}{RGB}{87,132,89}
\definecolor{muted-red}{RGB}{218,90,84}

\colorlet{cgreen}{muted-green}
\colorlet{cred}{muted-red}

\colorlet{my-blue}{my-full-blue!30}
\colorlet{my-orange}{my-full-orange!30}
\colorlet{my-green}{my-full-green!30}
\colorlet{my-red}{my-full-red!30}
\colorlet{my-purple}{my-full-purple!30}
\colorlet{my-brown}{my-full-brown!30}
\colorlet{my-pink}{my-full-pink!30}
\colorlet{my-gray}{my-full-gray!30}
\colorlet{my-olive}{my-full-olive!30}
\colorlet{my-cyan}{my-full-cyan!30}

%% file: headers/lqa/listing.tex
\usepackage{listings}

\usepackage{textcomp}

\usepackage{xcolor}

\usepackage[scaled=0.8]{beramono}

\definecolor{ckeyword}{HTML}{7F0055}
\definecolor{ccomment}{HTML}{3F7F5F}
\definecolor{cstring}{HTML}{2A0099}

\lstdefinestyle{numbers}{
	numbers=left,
	framexleftmargin=20pt,
	numberstyle=\tiny,
	firstnumber=auto,
	numbersep=1em,
	xleftmargin=2em
}

\lstdefinestyle{layout}{
	frame=none,
	captionpos=b,
}

\lstdefinestyle{comment-style}{
	morecomment=[l]//,
	morecomment=[s]{/*}{*/},
	commentstyle={\color{ccomment}\itshape},
}

\lstdefinestyle{string-style}{
	morestring=[b]",%
	morestring=[b]',%
	stringstyle={\color{cstring}},
	showstringspaces=false,%
}

\lstdefinestyle{keyword-style}{
	keywordstyle={\ttfamily\bfseries},
	morekeywords={
		function,
		constructor,
		int,
		bool,
		return,
		returns,
		uint
	},
	morekeywords = [2]{},
	keywordstyle = [2]{\text},
	sensitive=true,
}

\lstdefinestyle{input-encoding}{
	inputencoding=utf8,
	extendedchars=true,
	literate=
	{ℝ}{$\reals$}1%
	{→}{$\rightarrow$}1%
	{α}{$\alpha$}1%
	{β}{$\beta$}1%
	{λ}{$\lambda$}1%
	{θ}{$\theta$}1%
	{ϕ}{$\phi$}1%
}

\lstdefinestyle{escaping}{
	moredelim={**[is][\color{blue}]{\%}{\%}},
	escapechar=|,
	mathescape=true
}

\lstdefinestyle{default-style}{
	basicstyle=\fontencoding{T1}\ttfamily\footnotesize,
	style=numbers,
	style=layout,
	style=comment-style,
	style=string-style,
	style=keyword-style,
	style=input-encoding,
	style=escaping,
	tabsize=2,
	upquote=true
}

\lstdefinelanguage{BASIC}{
	language=C++,
	style=default-style
}[keywords,comments,strings]%

%% file: headers/lqa/tikz.tex
\usepackage{tikz}

\usetikzlibrary{arrows}
\usetikzlibrary{automata}
\usetikzlibrary{calc}
\usetikzlibrary{backgrounds}
\usetikzlibrary{decorations.markings}
\usetikzlibrary{decorations.pathmorphing}
\usetikzlibrary{decorations.pathreplacing}
\usetikzlibrary{fit}
\usetikzlibrary{patterns}
\usetikzlibrary{positioning}
\usetikzlibrary{shadows}
\usetikzlibrary{shapes}
\usetikzlibrary{shapes.geometric}

%% file: headers/header.tex
\input{math_commands.tex}

\usepackage[hidelinks]{hyperref}
\usepackage{url}

\usepackage{adjustbox}
\usepackage[capitalise,noabbrev]{cleveref}
\usepackage{scrextend}

\usepackage{enumitem}
\usepackage{threeparttable}
\usepackage{multirow, makecell}
\usepackage{booktabs}
\usepackage{xspace}
\usepackage{wrapfig}
\usepackage{dsfont}
\usepackage[normalem]{ulem}

\usepackage{subcaption}
\usepackage{caption}
\usepackage{dashrule}
\usepackage{amsthm}
\usepackage{thmtools}
\usepackage{thm-restate}
\usepackage{mathtools}
\usepackage[makeroom]{cancel}

\declaretheoremstyle[
  spaceabove=0.5em, %
  spacebelow=0.01em,%
  headfont=\normalfont\bfseries,
  bodyfont=\normalfont,
  postheadspace=0.5em,
]{mystyle}

\declaretheoremstyle[
  spaceabove=0.5em, %
  spacebelow=0.01em,%
  headfont=\normalfont\itshape,
  bodyfont=\normalfont,
  postheadspace=0.5em,
  qed=$\square$,
]{myproofstyle}

\crefname{thm}{Theorem}{Theorems}

\crefname{lem}{Lemma}{Lemmas}

\crefname{prop}{Proposition}{Propositions}

\crefname{cor}{Corollary}{Corollaries}

\crefname{defi}{Definition}{Definitions}

\usepackage{tikz}
\usetikzlibrary{calc,decorations,decorations.pathmorphing}
\usetikzlibrary{positioning,fit,arrows}
\usetikzlibrary{decorations.markings}
\usetikzlibrary{shapes,shapes.geometric}
\usetikzlibrary{shadows,patterns,snakes}
\usetikzlibrary{backgrounds,decorations.pathreplacing,automata}
\usetikzlibrary{intersections}
\usetikzlibrary{angles,quotes}
\usetikzlibrary{plotmarks}
\usetikzlibrary{patterns}
\usetikzlibrary{arrows.meta}
\usetikzlibrary{shapes.misc}
\usetikzlibrary{chains}
\usepackage{siunitx}
\newcolumntype{d}[1]{S[table-format=#1]}
\usetikzlibrary{patterns.meta}

\makeatletter
\def\extractcoord#1#2#3{
	\path let \p1=(#3) in \pgfextra{
		\pgfmathsetmacro#1{\x{1}/\pgf@xx}
		\pgfmathsetmacro#2{\y{1}/\pgf@yy}
		\xdef#1{#1} \xdef#2{#2}
	};
}
\makeatother

\usepackage{pifont}%

\DeclareMathOperator*{\din}{{\mathit{d}_{\text{in}}}}
\DeclareMathOperator*{\dout}{{\mathit{d}_{\text{out}}}}

\newcommand{\cifar}{CIFAR-10\xspace}

\newcommand{\TIN}{\textsc{TinyImageNet}\xspace}

\newcommand{\mnist}{\textsc{MNIST}\xspace}

\usepackage{comment}

%% file: math_commands.tex
\usepackage{amsmath,amsfonts,bm}

\def\1{\bm{1}}

\def\vk{{\bm{k}}}

\def\vp{{\bm{p}}}
\def\vq{{\bm{q}}}

\def\vw{{\bm{w}}}
\def\vx{{\bm{x}}}

\def\mW{{\bm{W}}}

\def\mY{{\bm{Y}}}

\DeclareMathAlphabet{\mathsfit}{\encodingdefault}{\sfdefault}{m}{sl}
\SetMathAlphabet{\mathsfit}{bold}{\encodingdefault}{\sfdefault}{bx}{n}

\def\gC{{\mathcal{C}}}
\def\gD{{\mathcal{D}}}
\def\gE{{\mathcal{E}}}

\def\gN{{\mathcal{N}}}

\def\sN{{\mathbb{N}}}

\def\sR{{\mathbb{R}}}

\DeclareMathOperator*{\argmax}{arg\,max}

%% file: headers/header-lqa-tail.tex
\input{headers/lqa/references}

%% file: headers/lqa/references.tex

\crefformat{section}{\S#2#1#3}

\crefrangeformat{section}{\S#3#1#4\crefrangeconjunction\S#5#2#6}

\crefmultiformat{section}{\S#2#1#3}{\crefpairconjunction\S#2#1#3}{\crefmiddleconjunction\S#2#1#3}{\creflastconjunction\S#2#1#3}

\newcommand{\crefrangeconjunction}{--}

\crefname{equation}{Equation}{Equations}

\crefname{listing}{Lst.}{listings}
\crefname{line}{Lin.}{Lin.}
\crefname{appendix}{Appendix}{Appendices}

\newcommand{\appref}[1]{%
	\ifbool{includeappendix}{\cref{#1}}{the appendix}%
}
\newcommand{\Appref}[1]{%
	\ifbool{includeappendix}{\cref{#1}}{The appendix}%
}

%% file: abstract.tex
Floating-point neural networks dominate modern machine learning but incur substantial
inference costs, motivating emerging interest in Boolean networks for resource-constrained deployments. Since Boolean networks use only Boolean operations, they can achieve nanosecond-scale inference latency.
However, learning Boolean networks that are both compact and accurate remains challenging
because of their discrete, combinatorial structure. In this work we address this challenge via three novel, complementary contributions: (i) a new parameter-free strategy for learning effective connections, (ii) a novel compact convolutional Boolean architecture that exploits spatial locality while requiring fewer Boolean operations than existing convolutional kernels, %
and (iii) an adaptive discretization procedure that reduces the accuracy drop incurred when converting a 
continuously relaxed
network into a discrete Boolean network. 
Across standard vision benchmarks, our method improves the Pareto frontier over prior state-of-the-art methods, achieving higher accuracy with up to $47\times$ fewer Boolean operations. This advantage also extends to other modalities. Further, on an FPGA, our 
model on \mnist
achieves 99.38\% accuracy with 6.48 ns latency, surpassing the prior state-of-the-art in both accuracy and runtime, while generating a $7\times$ smaller circuit. Code and models are available at \url{https://github.com/eth-sri/CompactLogic}.

%% file: sections/introduction.tex
\section{Introduction}\label{sec:intro}

Despite the success of neural networks in various domains, their high inference cost
causes heavy burden on the environment and limits their deployment
in resource-constrained environments such as edge devices and IoT applications. 
A promising solution
is to learn Boolean networks (a special family of Boolean circuits, c.f. \cref{sec:background}) that operate directly on Boolean values, i.e., 0 and 1, instead of on floating-point numbers,
because 
Boolean operations are inherently more efficient in terms of computation, memory and power consumption. 

A smaller Boolean circuit is generally cheaper to execute. 
For a given Boolean function, finding a compact,
or even minimal, Boolean circuit implementing this function is 
a fundamental and well-studied problem \citep{rudellMultipleValuedMinimizationPLA1987, ganai2000fly, micheli1994synthesis, bjesseDAGawareCircuitCompression2004,
 mishchenkoDAGawareAIGRewriting2006, rienerBooleanRewritingStrikes2022}. 
However, in practice, the Boolean function
is often not known a priori and must be learned from data. 

\textbf{This Work: Learning Compact and Accurate Boolean Networks Efficiently. }
We build on the learning framework proposed by \citet{petersen2022deep}. It works by
first relaxing a Boolean network into a differentiable representation and training it with gradient-based methods, 
and then discretizing it back into a Boolean network.
Despite existing efforts, three key challenges remain open. 
(i) \emph{Efficient connection learning}. 
Existing methods either take 
randomly fixed connection 
 \citep{petersen2022deep,petersen2024convolutional, ruttgers2025light}
 (i.e., no learning at all)
or rely on parameterization with 
additional weight matrices for learning \citep{bacellar2024differentiable, fojcik2025lilogic}. 
 By the lottery ticket hypothesis \citep{frankle2018the}, the ultra sparsity of Boolean networks (two inputs per neuron)  makes random connection highly ineffective.
The weight matrix parameterization, on the other hand, induces significant parameter overhead during training, typically quadratic in the number of neurons. How to efficiently learn effective connections is open. 
(ii) \emph{Compact convolutional architecture}. The only existing 
convolutional Boolean architecture  \citep{petersen2024convolutional} 
utilizes  a hardcoded tree-like kernel, which requires much more Boolean operations (exponential in the depth of the tree) than a feedforward layer.
(iii) \emph{Discretization-aware training}. The discrepancy between the differentiable relaxation and the discrete Boolean network causes a clear accuracy drop after discretization.

\begin{figure}[t]
  \centering
  \includegraphics[width=.5\columnwidth]{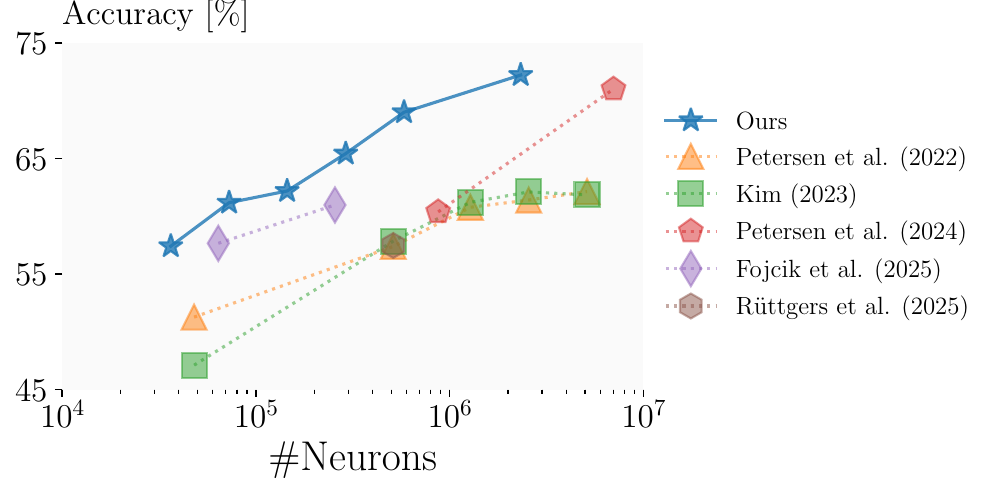} %
  \caption{Comparison between our method and prior works on \cifar. Note the log scale on x-axis.} \label{fig:acc_vs_gc}
  \vspace{-1em}
\end{figure}

In this paper, we address these three %
challenges 
with the overarching goal of 
learning
compact and accurate Boolean networks. Our main contributions are as follows.
\begin{itemize}[leftmargin=1.5em, noitemsep, topsep=-2pt]
   \item We introduce a parameter-free connection-learning strategy. By adaptively resampling candidate input-function triples, our method can learn effective sparse connectivity while avoiding the heavy matrix parameterization used in prior work. 
  \item We design a compact convolutional architecture that exploits spatial locality with single-operation Boolean kernels, 
  substantially reducing the size of 
  convolutional Boolean networks compared to prior work.
  \item We develop an adaptive discretization strategy that progressively discretizes network layers during training, narrowing the discretization gap and improving the performance of  the 
  final discrete model.
  \item Through 
  extensive evaluations, %
  our method substantially improves the Pareto frontier (\cref{fig:acc_vs_gc}), achieving higher accuracy with up to $47\times$ fewer Boolean operations. Additional results on tabular and sequence data, as well as FPGA synthesis on an edge device, further demonstrate the generality and inference-time efficiency of our approach.
\end{itemize}

%% file: sections/related_work.tex
\section{Related Work} \label{sec:related_work}
\textbf{Improving Inference Efficiency. } A large body of work 
attempts to reduce the inference cost of neural networks 
through quantization or pruning. 
Quantization reduces the computation cost through lowering 
the precision of weights and activations (e.g., to 8-bit, 4-bit, or even 1-bit with the so-called Binary neural networks),
while pruning reduces computation by removing connections or units. We refer to \citep{Gholami2021QuantSurvey, liu2025low} for surveys on quantization, and to \citep{cheng2024pruningsurvey} for a survey on pruning. 
In contrast, Boolean network represents a fundamentally different model than those obtained through 
quantization or pruning: they operate directly on the lowest computation level, namely Boolean, with no weights on the connecting edges. 
Note that Boolean networks are distinct from binary neural networks (c.f. \cref{app:current_effort_on_efficient_inference}).
\cite{petersen2022deep, petersen2024convolutional} report that Boolean networks can achieve much lower inference cost than current quantization or pruning methods.

\textbf{Learning Boolean Structures. } We review 
three families of learned Boolean structures: Boolean networks, binary decision diagrams (BDDs), and lookup tables (LUTs). \citet{petersen2022deep} introduced the differentiable learning framework for Boolean networks, followed  by
efforts in 
reducing training overhead \citep{ruttgers2025light}, learning inter-layer connections \citep{bacellar2024differentiable,fojcik2025lilogic},  convolutional design \citep{petersen2024convolutional}, and recurrent architectures \citep{buhrer2025recurrent}. For BDDs, exact learners formulate structure learning with SAT/MaxSAT \citep{Hu2022MaxSATBDD,Shati2023SATBDD} or mixed-integer programming \citep{Florio2023ODD}. \citet{Zantedeschi2021ArgminTrees} train discrete decision trees via argmin differentiation, and \citet{Qiu2025OBDDNet} parameterize ordered BDDs with continuous encodings before extracting discrete BDDs. For LUTs, prior work reduces sparse quantized networks into LUT networks \citep{Umuroglu2020LogicNetsFCCM}, trains LUT networks via Lagrange-interpolating polynomial relaxations \citep{Wang2019LUTNet}, incorporates hardware-aware training and assembly \citep{Andronic2025NeuraLUTAssembleFCCM}, performs post-training LUT compression \citep{Cassidy2025ReducedLUT}, and assembles small LUT subnetworks into larger LUT networks \citep{Weng2025AmigoLUT}.
A summary of the methods mentioned above on reducing the inference cost, along with their performance, can be found in 
\cref{app:current_effort_on_efficient_inference}.

%% file: sections/background.tex
\section{Background} \label{sec:background}

\textbf{Boolean Circuits and Boolean Networks. } Boolean circuits are 
directed acyclic graphs where each node computes a bivariate 
Boolean function $B:\{0,1\}^2\rightarrow \{0,1\}$.
We denote the $16$ possible bivariate Boolean functions by 
$B_i: \{0,1\}^2 \rightarrow \{0,1\}, i=1,\ldots, 16$ (c.f. \cref{app:bool_fun}).
A Boolean network is a Boolean circuit 
where nodes and edges are arranged in layers. \cref{fig:boolean_circuit_and_network} illustrates a Boolean circuit and a Boolean network.

\begin{figure}
\centering
\resizebox{.5\linewidth}{!}{\input{figures/3-1.tex}}
\caption{Left: a Boolean circuit. Right: a Boolean network.}\label{fig:boolean_circuit_and_network}
\vspace{-1em}
\end{figure}

Boolean circuits realize only Boolean functions.
To apply them to implement a non-Boolean function $g:A\rightarrow B$, 
we need to decompose $g$ into a Boolean function $f:\{0,1\}^{m} \rightarrow \{0,1\}^n, m,n\in \sN$, together with 
an encoder $\gE:A \rightarrow \{0,1\}^m$ and a decoder $\gD: \{0,1\}^n\rightarrow B$, such that $g=\gD  \circ f\circ \gE$.
In particular, for an image classification function 
  $g:[0,1]^d \rightarrow \gC$  that takes a real-valued image and maps it to a class label, we %
  use the thermometer encoder, defined as $\gE(\vx) := [\mathds{1}\{\vx \ge \frac{1}{N}\}, \dots, \mathds{1}\{\vx \ge \frac{N-1}{N}\}]$, where $\mathds{1}\{\cdot\}$ is the indicator function. We use the population count decoder (also named GroupSum) selecting the class with
 the highest number of 1s in the output Boolean vector $\mY \in \{0,1\}^{|\gC| \times n_c}$, defined as $\gD(\mY):= \argmax_{c \in \gC} \sum_{i=1}^{n_c} \mY_{c,i}$.

\textbf{Learning Boolean Networks via Differentiable Relaxations. } To learn a Boolean network
 from data, \citet{petersen2022deep} propose to 
relax  discrete Boolean networks into differentiable functions, enabling training via gradient descent.
Specifically, for $i=1,\ldots, 16$, let $B_i^c:[0,1]^2\rightarrow [0,1]$ be a differentiable relaxation of $B_i$. 
For example, $B_2(x_b,y_b) = x_b\wedge y_b, x_b,y_b\in \{0,1\}$, is relaxed into $B_2^c(x,y) = xy, x,y\in [0,1]$, satisfying $B_2^c(x_b, y_b) = B_2(x_b, y_b)$.
Each neuron is then parametrized as a weighted aggregation of all $B_i^c$ as 
\vspace{-0.2cm}
\begin{equation} \label{eq:relaxation}
    f^c(x,y) = \sum_{i=1}^{16} \frac{\exp(\vw_i)}{\sum_{j=1}^{16} \exp(\vw_j)} B_i^c(x,y),
\end{equation}
where $\vw_i$ are the parameters.
Afterwards, the relaxed network is trained on data by updating the weight vector $\vw = (\vw_1,\ldots,\vw_{16})$ of each neuron. 
Finally, each neuron is discretized back to a bivariate Boolean function by 
\vspace{-0.2cm}
\begin{equation} \label{eq:discretization}
    f := B_{i^*},  \quad \text{ with } i^* = \argmax_{i} \vw_i.
\end{equation}
Note that the relaxation and parametrization above only allow
to learn the Boolean operation at each neuron, but does not allow the learning of network connections. 
We shall address the efficient connection learning  later in \cref{sec:method}.

\textbf{Convolutional Boolean Networks. } Convolutional layers are key components 
in floating-point neural networks 
capturing local spatial structures. 
\citet{petersen2024convolutional} design a convolutional architecture for Boolean networks,  where each kernel is a binary tree consisting of Boolean operations. 
A tree of depth $d$ requires $2^d-1$ Boolean operations to 
implement a Boolean function with $2^{d-1}$ inputs. As a result, their convolutional layer 
costs significantly more Boolean operations than a non-convolutional layer.

%% file: figures/3-1.tex
\begin{tikzpicture}[
  gate/.style={circle, draw, minimum size=4mm, inner sep=0pt, font=\small},
  io/.style={font=\small},
  wire/.style={-}
]

\node[io] (a) at (1,0) {$x_1$};
\node[io] (b) at (1,1) {$x_2$};
\node[io] (c) at (1,2) {$x_3$};
\node[io] (d) at (6,0.5) {};
\node[io] (e) at (6,1.5) {};
\node[gate] (g1) at (2,2) {$B_1$};
\node[gate] (g2) at (3,1) {$B_4$};
\node[gate] (g3) at (5,0.5) {$B_7$};
\node[gate] (g4) at (5,1.5) {$B_3$};

\draw[-Stealth] (c.east) -- (g1.west);
\draw[-Stealth] (b.east) -- (g1.west);
\draw[-Stealth] (g1.east) -- (g2.west);
\draw[-Stealth] (a.east) -- (g2.west);
\draw[-Stealth] (a.east) -- (g3.west);
\draw[-Stealth] (g1.east) -- (g3.west);
\draw[-Stealth] (g3.east) -- (d.west);

\draw[-Stealth] (g2.east) -- (g4.west);
\draw[-Stealth] (g1.east) -- (g4.west);

\draw[-Stealth] (g4.east) -- (e.west);

\begin{scope}[xshift=7cm]

\node[io] (a) at (1,0) {$x_1$};
\node[io] (b) at (1,1) {$x_2$};
\node[io] (c) at (1,2) {$x_3$};
\node[io] (o1) at (4,1) {};
\node[io] (o2) at (4,2) {};
\node[io] (o3) at (4,0) {};

\node[gate] (g11) at (2,1) {$B_1$};
\node[gate] (g12) at (2,2) {$B_2$};
\node[gate] (g13) at (2,0) {$B_5$};
\node[gate] (g21) at (3,1) {$B_7$};
\node[gate] (g22) at (3,2) {$B_2$};
\node[gate] (g23) at (3,0) {$B_4$};

\draw[-Stealth] (a.east) -- (g11.west);
\draw[-Stealth] (b.east) -- (g11.west);
\draw[-Stealth] (b.east) -- (g12.west);
\draw[-Stealth] (c.east) -- (g12.west);
\draw[-Stealth] (c.east) -- (g13.west);
\draw[-Stealth] (a.east) -- (g13.west);

\draw[-Stealth] (g12.east) -- (g21.west);
\draw[-Stealth] (g13.east) -- (g21.west);
\draw[-Stealth] (g12.east) -- (g22.west);
\draw[-Stealth] (g13.east) -- (g22.west);
\draw[-Stealth] (g11.east) -- (g23.west);
\draw[-Stealth] (g13.east) -- (g23.west);
\draw[-Stealth] (g23.east) -- (o3.west);
\draw[-Stealth] (g22.east) -- (o2.west);
\draw[-Stealth] (g21.east) -- (o1.west);

\end{scope}
\end{tikzpicture}

%% file: sections/method.tex
\section{Learning Compact Boolean Networks} \label{sec:method}

In this section, we introduce our  methods to learn compact and accurate Boolean networks. 
We first present an efficient connection learning strategy in \cref{sec:learning_connectivity}, which is based on a novel parameterization of Boolean neurons
 that aggregates information from multiple connections, and an adaptive resampling strategy to continuously explore new connections.
Next, we present a compact convolutional Boolean architecture in \cref{sec:conv_without_trees}, replacing tree structures with single-operation
 kernels enabled by connection learning.
Finally, we introduce an adaptive discretization strategy in \cref{sec:adaptive_discretization}, progressively discretizing network layers during training to
reduce the accuracy drop.
\begin{figure*}[t]
\noindent\hbox to \textwidth{%
  \hfil
  \begin{subfigure}[t]{0.23\textwidth}
    \centering
    \resizebox{.7\linewidth}{!}{%
      \input{figures/single_input_gate.tikz}
    }
    \caption{\tiny Fixed connection. Each neuron observes two inputs.}
    \label{fig:conn-a}
  \end{subfigure}
  \hfil
  \begin{subfigure}[t]{0.23\textwidth}
    \centering
    \resizebox{.7\linewidth}{!}{%
      \input{figures/multi_input_gate.tikz}
    }
    \caption{\tiny Proposed parameterization. Each neuron observes multiple input pairs.}
    \label{fig:conn-b}
  \end{subfigure}
  \hfil
  \begin{subfigure}[t]{0.23\textwidth}
    \centering
    \resizebox{.7\linewidth}{!}{%
      \input{figures/gate_sum.tikz}
    }
    \caption{\tiny Aggregation in fixed connection (up) and the proposed method (down).}
    \label{fig:conn-c}
  \end{subfigure}
  \hfil
  \begin{subfigure}[t]{0.23\textwidth}
    \centering
    \resizebox{.7\linewidth}{!}{%
      \input{figures/discrete_gate.tikz}
    }
    \caption{\tiny After discretization, each neuron is a single Boolean operation.}
    \label{fig:conn-d}
  \end{subfigure}
  \hfil
}
\caption{An example comparing the parameterization with fixed connections (literature) and the proposed parameterization.  $B_?$ denotes undecided Boolean function.}
\label{fig:conn-all}
\end{figure*}

\subsection{Efficient Connection Learning} \label{sec:learning_connectivity}
We modify the parameterization in \cref{eq:relaxation} 
to allow connection learning as follows. Instead of 
parameterizing neurons as a weighted aggregation 
of sixteen Boolean operations with the same inputs, 
we assign different inputs to different Boolean operations.
Specifically, 
let $\vx \in [0,1]^d$ be the vector representing the $d$ neurons in the $\ell$-th layer. 
For a neuron in the $(\ell+1)$-th layer, we parameterize it as  
\vspace{-0.2cm}
\begin{equation} %
  \tilde{f^c}(\vx; \vk, \vp, \vq) = \sum_{i=1}^{16} \frac{\exp(\vw_i)}{\sum_{j=1}^{16} \exp(\vw_j)} B_{\vk_i}^c(\vx_{\vp_i}, \vx_{\vq_i}),
\end{equation}
where $\vk \in \{1, \dots, 16\}^{16}$ are indices denoting the Boolean operations, and $\vp, \vq \in \{1,\dots, d\}^{16}$
are indices denoting the two input neurons from the $\ell$-th layer, respectively.
At the beginning of training, $(\vk, \vp, \vq)$ can be sampled uniformly at random or follow any specific assignments. See \cref{fig:conn-all} for an illustration of the  parameterization.

For each neuron, the optimal combination of inputs and Boolean function might not be included
in the initial assignment. To learn better combinations, we propose an adaptive 
resampling strategy.
The high-level idea is: once a neuron becomes ``stable'' during training, 
we resample its $(\vk, \vp, \vq)$ candidates to explore new combinations.
We measure the stability of a neuron by monitoring 
its  weight entropy, defined as:
\vspace{-0.2cm}
\begin{equation*}
  h(\vw) = - \sum_{i=1}^{16} \tilde{\vw}_i \log \tilde{\vw}_i, \text{ where } \tilde{\vw}_i = \frac{\exp(\vw_i)}{\sum_{j=1}^{16} \exp(\vw_j)}.
\end{equation*}
The weight entropy equals zero if and only if one element in the normalized weight $\tilde{\vw}$ equals one, indicating that the neuron is represented by a single $(\vk_i, \vp_i, \vq_i)$ triple, and reaches its maximum when all elements in $\tilde{\vw}$ are equal, indicating that the neuron is an equal mixture of all $(\vk_i, \vp_i, \vq_i)$ triples.
We track the exponential moving average (EMA) of the weight entropy 
for each neuron individually.
 When the EMA of $h(\vw)$ for a neuron 
 converges, its weight vector $\vw$ can have three possible states:
(i) dominated by a single triple $(\vk_i, \vp_i, \vq_i)$, i.e., $\exists i^*, \tilde{\vw}_{i^*} \ge 0.95$,
(ii) dispersed, i.e., $\max_i \vw_i \le 0.4$, or
(iii) concentrated without a dominant triple. 
For (i), further training of the neuron is unlikely to improve the discretized network unless the dominant triple is replaced, so we resample all $(\vk_i, \vp_i, \vq_i)$ except the dominant one to explore new combinations.
For (ii), the neuron fails to find a useful triple, thus we resample all $(\vk_i, \vp_i, \vq_i)$ to restart its learning. Neurons falling into the third case are not changed since they are still learning. A full pseudocode of the resampling strategy is provided in \cref{alg:resampling} in \cref{app:pseudocode}.
We note that after training, most neurons (frequently over 90\%)
 empirically end up in the first case, indicating that they have learned meaningful patterns expressed by a single dominant connection and Boolean function.

We remark on the relation to prior works and the overhead of the proposed method.
As mentioned in \cref{sec:intro}, 
\citet{petersen2022deep,petersen2024convolutional} and \citet{ruttgers2025light}
randomly sample the connection and keep it fixed during training. 
 To allow learning connections,  \citet{bacellar2024differentiable} and \citet{fojcik2025lilogic} 
propose to parameterize the connections with additional weight matrices.
A Boolean layer with $\din$ input neurons and $\dout$ 
output neurons additionally maintains a weight matrix $W_{\text{link}} \in \sR^{\din \times \dout}$.
However, such large matrices are 
 often impractical: for a typical medium-sized model in  \citet{petersen2024convolutional}, $W_{\text{link}}$ for a single 
 layer can consume over 3400 GB of memory. In contrast, our method learns the network connection without 
 additional 
 parameters.

\subsection{Convolution without Boolean Trees} \label{sec:conv_without_trees}
Convolution is a key component in floating-point neural networks 
to capture spatial features, especially for vision tasks.
A natural and easy way to construct convolutional layers in Boolean networks 
is to take a single Boolean operation
 as the convolutional kernel. 
However, such kernels under fixed connections only observe two inputs from 
the receptive field.   
With the design in  \cref{sec:learning_connectivity},  this does not constitute a problem:
 a convolutional kernel 
  learns to extract important information from the full receptive field by aggregating 
  multiple input pairs during training, thus the kernel can observe up to 32 distinct inputs from the receptive field. As a result, we can learn a compact convolutional Boolean network based on single-operation kernels.

  In contrast, the convolutional kernels in \citet{petersen2024convolutional}
are built on Boolean trees.  Specifically, $2^{d-1}$ neurons sampled 
from the receptive field are fed into a binary Boolean tree of depth $d$, and then every node in the tree 
is learned with the relaxation in \cref{eq:relaxation}. Such tree structure has two main issues. First,
 the number of Boolean operations per layer grows exponentially with the tree depth $d$, leading to high inference costs.
  Second, the hierarchical structure of binary trees introduces sequential dependencies that limit parallelism during both 
  forward and backward passes. As we later demonstrate empirically in \cref{sec:experiments},
  our method can learn much smaller convolutional Boolean networks with higher accuracies.

\subsection{Adaptive Discretization} \label{sec:adaptive_discretization}

\begin{figure}
    \centering
    \includegraphics[width=0.5\textwidth]{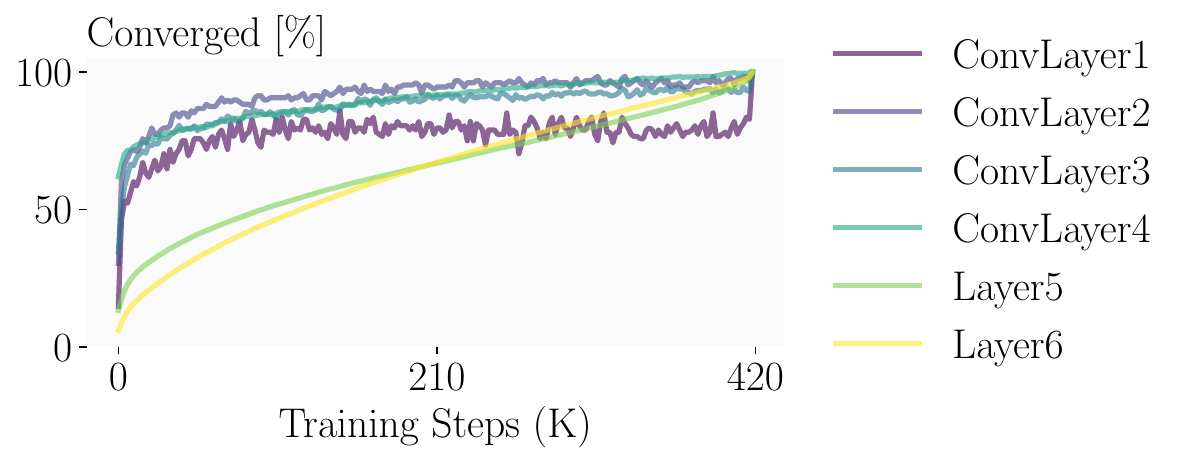}
    \caption{Convergence speed of different layers.}
    \label{fig:row_argmax_difference}
    \vspace{-1em}
\end{figure}

As introduced in \cref{sec:background}, after training, the network 
is discretized to Boolean by replacing all relaxed neurons with their dominant Boolean operation at once.  
However, such a training paradigm ignores the discrepancy 
between the continuous and Boolean networks, and thus leads to a systematic accuracy drop after discretization.
We shall address the discretization issue in this subsection. 

During training, convolutional layers of the network tend to converge faster than the non-convolutional layers. We measure the ratio of neurons that have the same dominant $(\vk_i, \vp_i, \vq_i)$ triple as the final discretized network. As shown in \cref{fig:row_argmax_difference}, convolutional layers converge much faster than non-convolutional layers. Further, convolutional layers tend to be less stable, \eg the first convolutional layer stops improving when roughly 80\% neurons converge, while non-convolutional layers converge more steadily.
Based on these observations, 
we propose to adaptively discretize (and thus freeze) the network progressively from shallow to deep layers during training.
 Specifically, we monitor the convergence of each layer using the average weight entropy and EMAs, similar to \cref{sec:learning_connectivity}.
Once the first layer converges, we immediately discretize it by \cref{eq:discretization}  (thus stop training it) and continue training the
subsequent layers. We proceed layer-wise in this manner for all convolutional layers, and then continue training the
non-convolutional layers in their relaxed form until the end.
A full pseudocode of the adaptive discretization strategy is provided in \cref{alg:discretization} in \cref{app:pseudocode}.
The progressive discretization  forces the layers to adapt to Boolean inputs during training. As a result,   the network suffers less from the accuracy drop caused by discretization.

We remark that
 \citet{kim2023deep} and \citet{yousefi2025mind} attempt to
 mitigate the discretization issue as well, but via injecting noise during training.
However, the injected noise distribution does not accurately reflect the true distribution of the discretization error, and thus fails to guide the training towards better solutions (c.f. \cref{sec:exp_discretization}).

%% file: figures/single_input_gate.tikz
\begin{tikzpicture}[
  x=1cm,y=1cm,
  neuron/.style={circle,draw=gray!70,fill=white,minimum size=5mm,inner sep=0pt,line width=0.6pt},
  bitbox/.style={draw=gray!70,fill=white,minimum size=5mm,inner sep=0pt,line width=0.6pt,font=\small},
  worange/.style={orange!90!black, line width=0.9pt, dashed, line cap=round},
  wblue/.style={blue!60!black, line width=0.9pt, dashed, line cap=round},
  wgreen/.style={green!60!black, line width=0.9pt, dashed, line cap=round},
  lab/.style={font=\scriptsize}
]

\def\xin{0}
\def\xhA{2.5}
\def\xout{5.0}

\foreach \i/\y/\b in {1/2.5/1,2/2.0/0,3/1.5/0,4/1.0/1,5/0.5/0,6/0.0/1}{
  \node[bitbox] (in\i) at (\xin,\y) {\b};
}

\foreach \i/\y in {1/2.5,2/1.25,3/0.0}{
  \node[neuron] (hA\i) at (\xhA,\y) {$B_?$};
}

\draw[worange] (in1.east) -- (hA1);
\draw[worange] (in2.east) -- (hA1);
\draw[worange] (in3.east) -- (hA2);
\draw[worange] (in4.east) -- (hA2);
\draw[worange] (in5.east) -- (hA3);
\draw[worange] (in6.east) -- (hA3);

\end{tikzpicture}

%% file: figures/multi_input_gate.tikz
\begin{tikzpicture}[
  x=1cm,y=1cm,
  neuron/.style={circle,draw=gray!70,fill=white,minimum size=5mm,inner sep=0pt,line width=0.6pt},
  bitbox/.style={draw=gray!70,fill=white,minimum size=5mm,inner sep=0pt,line width=0.6pt,font=\small},
  worange/.style={orange!90!black, line width=0.9pt, dashed, line cap=round},
  wblue/.style={blue!60!black, line width=0.9pt, dashed, line cap=round},
  wgreen/.style={green!60!black, line width=0.9pt, dashed, line cap=round},
  lab/.style={font=\scriptsize}
]

\def\xin{0}
\def\xhA{2.5}
\def\xout{5.0}

\foreach \i/\y/\b in {1/2.5/1,2/2.0/0,3/1.5/0,4/1.0/1,5/0.5/0,6/0.0/1}{
  \node[bitbox] (in\i) at (\xin,\y) {\b};
}

\foreach \i/\y in {1/2.5,2/1.25,3/0.0}{
  \node[neuron] (hA\i) at (\xhA,\y) {$B_?$};
}

\draw[worange] (in1.east) -- (hA1);
\draw[worange] (in2.east) -- (hA1);
\draw[wgreen] (in3.east) -- (hA1);
\draw[wgreen] (in4.east) -- (hA1);
\draw[wblue] (in5.east) -- (hA1);
\draw[wblue] (in6.east) -- (hA1);
\draw[wgreen] (in1.east) -- (hA2);
\draw[wgreen] (in2.east) -- (hA2);
\draw[worange] (in3.east) -- (hA2);
\draw[worange] (in4.east) -- (hA2);
\draw[wblue] (in5.east) -- (hA2);
\draw[wblue] (in6.east) -- (hA2);
\draw[wblue] (in1.east) -- (hA3);
\draw[wblue] (in2.east) -- (hA3);
\draw[worange] (in3.east) -- (hA3);
\draw[worange] (in4.east) -- (hA3);
\draw[wgreen] (in5.east) -- (hA3);
\draw[wgreen] (in6.east) -- (hA3);

\end{tikzpicture}

%% file: figures/gate_sum.tikz
\begin{tikzpicture}[
  font=\large,
  gate/.style={draw=#1, line width=1.2pt, fill=none, scale=1.25},
  plus/.style={font=\Large},
  w/.style={font=\large},
]
\colorlet{myorange}{orange!90!black}
\colorlet{mygreen}{green!60!black}
\colorlet{myblue}{blue!60!black}

\def\dx{-0.8}
\def\yA{3}
\def\yB{0}

\coordinate (r1a) at (0,\yA);
\coordinate (r1b) at (2.5,\yA);
\coordinate (r1c) at (5,\yA);

\node[nand gate US, logic gate inputs=nn, gate=myorange] (A1) at (r1a) {$B_3$};
\node[xor  gate US, logic gate inputs=nn, gate=myorange] (B1) at (r1b) {$B_7$};
\node[or   gate US, logic gate inputs=nn, gate=myorange] (C1) at (r1c) {$B_9$};

\node[plus] at ($(A1)!0.5!(B1) + (0,0.05)$) {$+$};
\node[plus] at ($(B1)!0.5!(C1) + (0,0.05)$) {$+$};

\node[w, text=myorange] at ($(A1)+(0,\dx)$) {0.7};
\node[w, text=myorange] at ($(B1)+(0,\dx)$) {0.2};
\node[w, text=myorange] at ($(C1)+(0,\dx)$) {0.1};

\coordinate (r2a) at (0,\yB);
\coordinate (r2b) at (2.5,\yB);
\coordinate (r2c) at (5,\yB);

\node[nand gate US, logic gate inputs=nn, gate=myorange] (A2) at (r2a) {$B_3$};
\node[xor  gate US, logic gate inputs=nn, gate=myblue]   (B2) at (r2b) {$B_7$};
\node[or   gate US, logic gate inputs=nn, gate=mygreen]  (C2) at (r2c) {$B_9$};

\node[plus] at ($(A2)!0.5!(B2) + (0,0.05)$) {$+$};
\node[plus] at ($(B2)!0.5!(C2) + (0,0.05)$) {$+$};

\node[w, text=myorange] at ($(A2)+(0,\dx)$) {0.7};
\node[w, text=myblue]   at ($(B2)+(0,\dx)$) {0.2};
\node[w, text=mygreen]  at ($(C2)+(0,\dx)$) {0.1};

\end{tikzpicture}

%% file: figures/discrete_gate.tikz
\begin{tikzpicture}[
  x=1cm,y=1cm,
  wire/.style={draw=gray!65,line width=0.6pt},
  bitbox/.style={draw=gray!70,fill=white,minimum size=5mm,inner sep=0pt,line width=0.6pt,font=\small},
  lbl/.style={font=\scriptsize}
]

\def\xin{0}
\foreach \i/\y/\b in {1/2.5/1,2/2.0/0,3/1.5/0,4/1.0/1,5/0.5/0,6/0.0/1}{
  \node[bitbox] (in\i) at (\xin,\y) {\b};
}

\def\xgate{2.5}
\node[nand gate US, draw=gray!70, logic gate inputs=nn, scale=0.85] (g1) at (\xgate,2.5) {$B_3$};
\node[xor gate US, draw=gray!70, logic gate inputs=nn, scale=0.85] (g2) at (\xgate,1.25) {$B_7$};
\node[or gate US, draw=gray!70, logic gate inputs=nn, scale=0.85] (g3) at (\xgate,0.0) {$B_9$};

\draw[wire] (in5.east) -- (g1.input 1);
\draw[wire] (in6.east) -- (g1.input 2);

\draw[wire] (in1.east) -- (g2.input 1);
\draw[wire] (in2.east) -- (g2.input 2);

\draw[wire] (in3.east) -- (g3.input 1);
\draw[wire] (in4.east) -- (g3.input 2);

\end{tikzpicture}

%% file: sections/experiment.tex
\section{Experiments} \label{sec:experiments}

We evaluate the proposed components both in isolation and in combination. Details on input encoding, data augmentation, baselines, architectures, and training hyperparameters are provided in \cref{app:exp_details}. We first evaluate learned connections (\cref{sec:exp_connection}), compact convolutions (\cref{sec:exp_convolution}), and adaptive discretization (\cref{sec:exp_discretization}) in isolation,
then 
 combine all components to evaluate the final Pareto frontier and FPGA inference performance (\cref{sec:exp_main}).

\textbf{Metrics. } We report test accuracy 
as the prediction-performance metric. To measure Boolean circuit size before hardware synthesis, we report both the number of neurons in the Boolean network (\emph{\#neurons}) and the number of Boolean operations after simple pruning (\emph{\#BOPs}). %
The simple pruning
removes redundant neurons that do not contribute to the final output, as well as Identity and NOT neurons by rewiring their input connections to their output connections. See \cref{app:pruning_algorithm} for details. We apply the same pruning procedure to all methods to avoid bias toward any particular circuit optimization method. For FPGA inference, we generate Verilog from the pruned Boolean networks and synthesize it with Xilinx Vivado for the ZCU104 FPGA. 
The synthesis 
performs more advanced hardware-level simplification and mapping than our simple pruning, so we report the resulting post-synthesis circuit size in MB and runtime per sample 
in nano-second. 

\subsection{Connection Learning} \label{sec:exp_connection}

We evaluate the connection-learning method from \cref{sec:learning_connectivity} against DiffLogicNet \citep{petersen2022deep} under matched architecture and data processing settings, differing only in the neuron parameterization.

\textbf{Main Results. } \Cref{tb:DLGN_CIFAR} summarizes the results on \cifar. 
Across all model sizes, learned connections as in our model consistently outperform randomly sampled fixed connections as in DiffLogicNet. Notably, our 512K-neuron model is more accurate than a DiffLogicNet having 1.28M neurons. On average, connection learning improves \cifar accuracy by 2.73\%. Results on \mnist are consistent, as reported in \cref{app:additional_ablation}.

\input{tables/DLGN.tex}

\begin{figure}
\centering
\begin{minipage}{.48\textwidth}
  \centering
  \includegraphics[height=2.3cm]{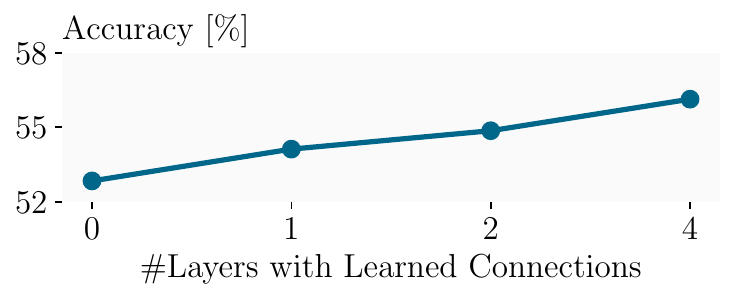}
  \captionof{figure}{Effect of the number of learned-connection \\ layers on \cifar. Note that the x-axis is unequ- \\ally  spaced.}
\label{fig:mip_ablaition}
\end{minipage}%
\begin{minipage}{.48\textwidth}
  \centering
  \includegraphics[height=2.2cm]{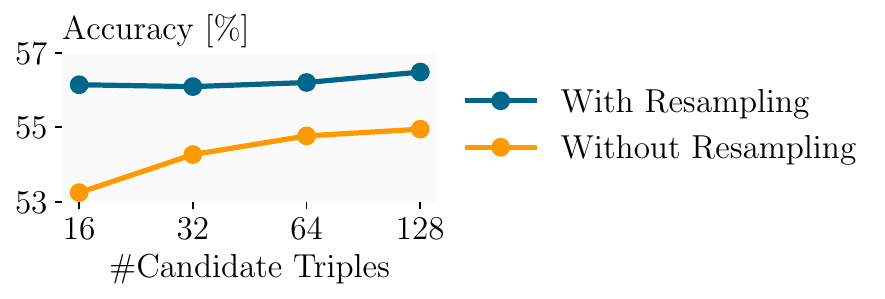}
  \captionof{figure}{The effect of increasing the number of candidate triples, with and without resampling. Note the log scale on x-axis.}
  \label{fig:acc_vs_k}
\end{minipage}
\end{figure}

\textbf{Learning More Layers Helps. } \citet{bacellar2024differentiable} reported that, 
for their connection-learning algorithm, learning only the first-layer connections can match fully learnable models, suggesting limited effectiveness of connection learning in deeper layers. 
We test whether our method learns useful connections throughout the network by varying the number of learned-connection layers in the small \cifar model, while keeping fixed random connections in the remaining layers as in DiffLogicNet. As shown in \cref{fig:mip_ablaition}, accuracy consistently improves as more layers learn connections, indicating that our method is effective in both shallow and deep layers.

\textbf{Resampling vs More Candidates. } As discussed in \cref{sec:learning_connectivity}, 
resampling explores new candidate triples during training. A direct alternative is to increase the number of candidate triples $(\vk_i, \vp_i, \vq_i)$ per neuron from 16 to a larger value $K$, allowing each neuron to aggregate information from a larger pool. As shown in \cref{fig:acc_vs_k}, increasing $K$ from 16 to 128 significantly improves accuracy when resampling is disabled, from 53.24\% to 54.94\%. With resampling enabled, however, the gain is marginal, from 56.14\% to 56.48\%. Further, resampling consistently outperforms the non-resampling counterpart for all $K$. Thus, resampling effectively learns good connections, making the parameter overhead of maintaining many candidates per neuron unnecessary.

\subsection{Single-Operation Boolean Convolution} \label{sec:exp_convolution}

We evaluate the proposed convolutional architecture, which uses a single Boolean operation as the convolutional kernel, against TreeLogicNet \citep{petersen2024convolutional}, which uses Boolean-tree kernels. Both architectures use $3\times3$ receptive fields. TreeLogicNet restricts each kernel to two neighboring channels, while our final models restrict each kernel to one channel. We show that this restriction improves performance, and we hypothesize that the effect arises from thermometer encoding on natural images.

\textbf{Main Results. } \Cref{tb:CLGN_CIFAR} summarizes the results on \cifar. Our architecture consistently outperforms TreeLogicNet with fewer neurons: our large model achieves 0.72\% higher accuracy (72.09\% vs. 71.37\%) with 3x fewer neurons (2.33M vs. 7M). We exclude the larger models from \citep{petersen2024convolutional} because they include floating-point feature-extraction layers and are therefore not Boolean networks. Results on \mnist show consistent improvements, 
as reported in detail in  \cref{app:additional_ablation}.

\input{tables/CLGN.tex}

\begin{figure}
    \centering
    \includegraphics[width=0.5\linewidth]{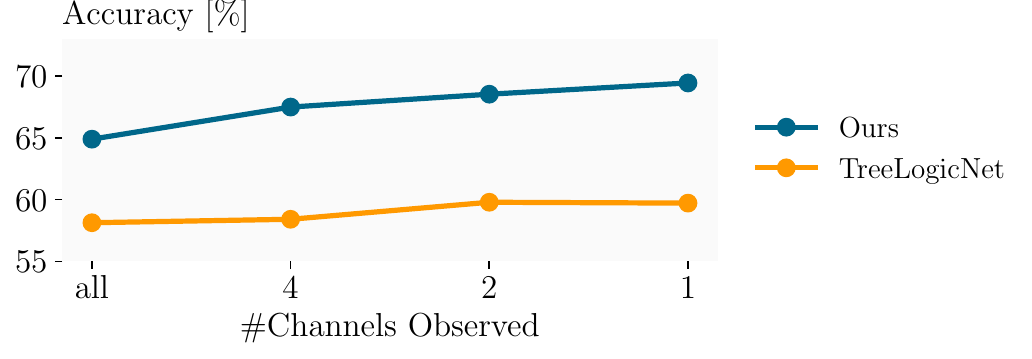}
    \caption{Comparison with TreeLogicNet under matched channel-visibility conditions. Across all channel visibilities, our method achieves higher accuracy while using fewer neurons.}
    \label{fig:matched_visibility_treelogic}
\end{figure}

\textbf{The Paradox of Channel Visibility. } To isolate the convolution contribution from channel visibility, we compare our method (582K model) and TreeLogicNet (874K model) under matched channel-visibility conditions. As shown in \cref{fig:matched_visibility_treelogic}, our smaller model consistently outperforms TreeLogicNet across all channel visibilities. This supports the advantage of connection-learned single-operation kernels over hard-coded Boolean-tree kernels. Since TreeLogicNet uses a hard-coded Boolean-tree kernel, applying our connection-learning strategy to it would require a substantial redesign. We therefore cannot further match connection learning across these two architectures.

This study reveals a paradox of channel visibility. Consistent with \citep{petersen2024convolutional}, limiting the number of accessible channels substantially improves performance on \cifar. In particular, restricting each kernel to one channel achieves the best accuracy, 4.56\% higher than using all channels (69.46\% vs. 64.90\%). This is surprising because one-channel visibility prevents convolutional layers from mixing information across channels. We hypothesize that the effect arises from thermometer encoding on natural images. As visualized in \cref{app:thermo_visual}, thermometer encoding can break complementary cross-channel information and sometimes removes information from some channels almost entirely. This may cause interference when a kernel aggregates multiple channels. An input encoding that preserves cross-channel information may therefore enable more effective channel mixing and further improve convolutional Boolean networks. To further test this hypothesis, we repeat the channel visibility experiments on \mnist in \cref{app:mnist_channel_visibility}, where thermometer encoding degenerates to one-hot encoding and thus does not break cross-channel information. The result shows that channel restriction does not improve performance on \mnist, which 
supports
the hypothesis.

\subsection{Adaptive Discretization} \label{sec:exp_discretization}

We evaluate adaptive discretization against Gumbel noise injection on both final test accuracy and the discretization gap, which is defined as the accuracy difference between the relaxed model before discretization and the discrete Boolean network after discretization, following \citet{yousefi2025mind}.

\begin{figure}
    \centering
    \includegraphics[width=0.48\textwidth]{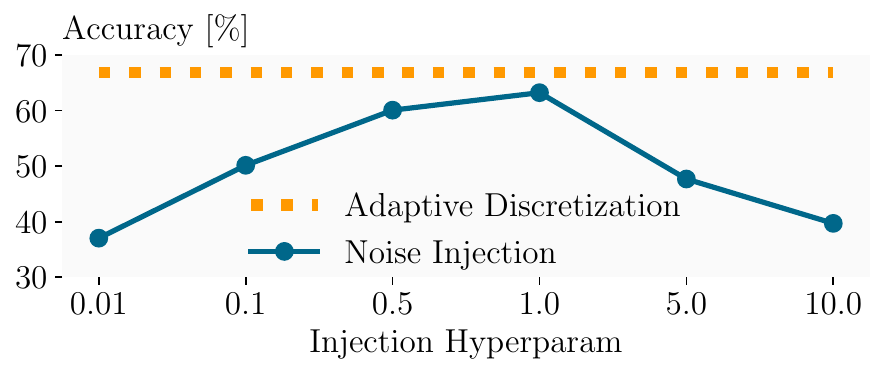}
    \caption{Comparison between Gumbel noise injection and the proposed adaptive discretization. Note that the x-axis is unequally spaced.}
    \label{fig:gumbel_vs_lh}
    \vspace{-.5cm}
\end{figure}

\textbf{Main Results. } \Cref{fig:gumbel_vs_lh} compares adaptive discretization with Gumbel noise injection on \cifar using our medium convolutional model. We sweep the Gumbel-noise hyperparameter for a fair comparison. Adaptive discretization consistently improves test accuracy while reducing training cost: after four convolutional layers are discretized, training is 1.8x faster than the fully relaxed model, since discretized layers are cheaper in both forward and backward passes. By contrast, Gumbel noise injection adds computational overhead throughout training. In addition, ablation studies in \cref{app:fixed_schedule_discretization} show that well-tuned fixed-schedule discretization could also match the performance of adaptive discretization, highlighting the importance of discretization-aware training. Nevertheless, adaptive discretization is more convenient in practice because it does not require tuning additional discretization schedules.

\begin{figure}
    \centering
    \includegraphics[width=0.48\textwidth]{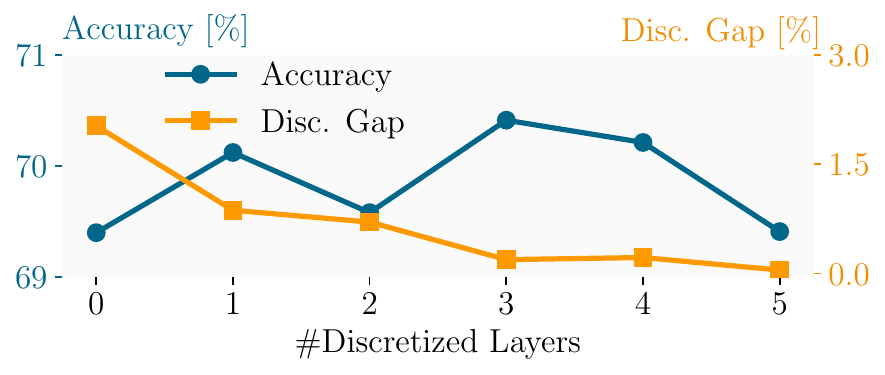}
    \caption{The effect of the number of discretized layers during training on the accuracy and discretization gap.}
    \label{fig:acc_vs_frozen_layers}
\end{figure}

\textbf{Progressive Discretization Reduces the Gap. } We further vary the number of layers discretized during training from 1 to 5 in the medium convolutional \cifar model, which has 6 layers in total. As shown in \cref{fig:acc_vs_frozen_layers}, discretizing more layers progressively reduces the discretization gap, confirming the intended effect of adaptive discretization. Discretizing all layers during training does not yield the best final accuracy under the fixed training budget, likely because early discretization limits the model's remaining learning capacity. In practice, we adaptively discretize all convolutional layers across all settings, balancing gap reduction with sufficient optimization, since non-convolutional layers already converge steadily without training-time discretization (c.f. \cref{fig:row_argmax_difference}).

\subsection{Combined Results and FPGA Inference} \label{sec:exp_main}
We combine all proposed components and evaluate the final models on \mnist and \cifar. \Cref{tb:exp_main} summarizes accuracy, Boolean-network complexity, and FPGA inference performance. On \mnist, our small and medium models achieve 99.31\% and 99.38\% accuracy, outperforming the large TreeLogicNet model with 45x and 22x fewer neurons (260K and 520K vs. 11.8M), respectively. After pruning, this corresponds to 47x and 24x fewer Boolean operations (130K and 251K vs. 6.2M). On \cifar, our large model reaches 72.56\% accuracy, outperforming TreeLogicNet with 3x fewer neurons and Boolean operations. Overall, the proposed components combine effectively to learn compact and accurate Boolean networks.

We further compile our models and TreeLogicNet models to Verilog and measure inference on a Xilinx ZCU104 FPGA. \Cref{tab:fpga_treelogic} shows that our method consistently reduces runtime relative to TreeLogicNet, while also reducing post-synthesis circuit size.

We include further results on \TIN, tabular data, and sequence data in \cref{app:tinyimagenet_results}, which show consistent improvements over prior methods, demonstrating the generality of our approach across dataset scale and data modalities.

\begin{table}[t]
\centering
\small
\caption{Final-model comparison with TreeLogicNet, including Boolean-network complexity and FPGA inference performance.}
\label{tab:merged}
\label{tb:exp_main}
\label{tab:fpga_treelogic}
\resizebox{0.98\columnwidth}{!}{
\begin{tabular}{ccccccc}
\toprule
Dataset & Method & Acc. (\%) & \emph{\#neurons} & \emph{\#BOPs} & FPGA Circuit (MB) & FPGA Runtime (ns) \\
\midrule

\multirow{5}{*}{\mnist}
& TreeLogicNet-S & 97.77 & 185 K  & 117 K  & 11.236  & 6.96 \\
& TreeLogicNet-M & 98.96 & 740 K  & 427 K  & 34.623  & 7.74 \\
& TreeLogicNet-L & 99.33 & 11.8 M & 6.20 M & 484.592 & 8.76 \\
\cmidrule(lr){2-7}
& Ours-S         & 99.31 & 260 K  & 130 K  & 32.424  & 6.18 \\
& Ours-M         & 99.38 & 520 K  & 251 K  & 63.599  & 6.48 \\

\midrule

\multirow{5}{*}{\cifar}
& TreeLogicNet-S & 59.79 & 874 K  & 521 K  & 49.265  & 8.16 \\
& TreeLogicNet-M & 71.37 & 7.00 M & 3.66 M & 337.524 & 9.06 \\
\cmidrule(lr){2-7}
& Ours-T         & 62.71 & 145 K  & 85.2 K & 19.667  & 5.88 \\
& Ours-S         & 67.12 & 291 K  & 163 K  & 38.339  & 6.18 \\
& Ours-M         & 70.22 & 582 K  & 319 K  & 76.406  & 6.48 \\
& Ours-L         & 72.56 & 2.33 M & 1.08 M & 266.926 & 7.08 \\

\bottomrule
\end{tabular}
}
\end{table}

\vspace{-1em}

%% file: tables/DLGN.tex
\begin{table}
\centering
\caption{Comparison between DiffLogicNet and our method on \cifar.}
\label{tb:DLGN_CIFAR}
  \resizebox*{.5\linewidth}{!}{%

\begin{tabular}{lccc}
\toprule
Method & Acc. & \emph{\#neurons} & \emph{\#BOPs} \\
\midrule
DiffLogicNet (small)    & $52.84\%$ & 48.0 K & 27.0 K \\  %
DiffLogicNet (medium)   & $59.90\%$ & 512 K & 312 K \\  %
DiffLogicNet (large)    & $62.57\%$ & 1.28 M & 780 K \\  %
\midrule
Ours (small)            & \textbf{$56.13\%$} & 48.0 K & 27.0 K \\  %
Ours (medium)           & \textbf{$62.84\%$} & 512 K & 297 K \\  %
Ours (large)            & \textbf{$64.53\%$} & 1.28 M & 705 K \\  %
\bottomrule
\end{tabular}
  }
\end{table}

%% file: tables/CLGN.tex
\begin{table}
\centering
\caption{Comparison between TreeLogicNet and our method on \cifar.}
\label{tb:CLGN_CIFAR}
  \resizebox*{.47\linewidth}{!}{%

\begin{tabular}{lccc}
\toprule
Method & Acc. & \emph{\#neurons} & \emph{\#BOPs} \\
\midrule
TreeLogicNet-S	& $59.79\%$ & 874 K & 521 K \\
TreeLogicNet-M	& $71.37\%$ & 7.00 M & 3.66 M \\
\midrule
Ours-S	& \textbf{$65.56\%$} & 291 K & 161 K \\
Ours-M	& \textbf{$68.73\%$} & 582 K & 307 K \\
Ours-L	& \textbf{$72.09\%$} & 2.33 M & 1.09 M \\
\bottomrule
\end{tabular}
  }
\end{table}

%% file: sections/discussion.tex
\section{Discussion} \label{sec:discussion}

Despite the promising results, several limitations are open for future work.
First, while our method effectively reduces the number of neurons during training and the number of Boolean operations at inference,
the training process still relies on floating-point computations, which can be resource-intensive for large models since each neuron contributes
16 floating-point weights.
Second, our training method does not take into account advanced logic synthesis techniques for network optimization.
Further exploiting certain properties of such methods to train synthesis-friendly networks
might lead to even more compact models after synthesis.
Finally, this work only considers the training process of the Boolean network. The design of input encoding and output decoding schemes
deserves further investigation.

%% file: sections/conclusion.tex
\section{Conclusion} \label{sec:conclusion}

This work presents a unified approach for learning compact and accurate Boolean networks. We address three bottlenecks in existing methods: inefficient random connection, expensive Boolean-tree convolutional kernels, and the discretization gap between relaxed and discrete networks. Our method learns sparse connections without additional parameterization matrices, %
replaces Boolean-tree kernels with single-operation convolutional kernels, and progressively discretizes layers during training. Across vision benchmarks, these designs improve accuracy while substantially reducing Boolean network complexity, achieving up to $47\times$ fewer Boolean operations than prior state-of-the-art methods. FPGA synthesis further shows that the resulting networks translate into lower-latency hardware implementations, and additional experiments on other modalities prove the generality of the proposed approach.

%% file: sections/appendix.tex
\section{Pseudocode for Adaptive Procedures} \label{app:pseudocode}

This appendix gives pseudocode for the two adaptive procedures introduced in \cref{sec:method}. \Cref{alg:resampling} describes adaptive resampling for connection learning (\cref{sec:learning_connectivity}), and \cref{alg:discretization} describes adaptive discretization during training (\cref{sec:adaptive_discretization}).

\begin{algorithm}[!h]
  \caption{Adaptive Resampling}
  \label{alg:resampling}
  \begin{algorithmic}
    \STATE {\bfseries Input:} initial neuron weight $\vw \in \sR^{16}$, weight update function $\STEP$, EMA momentum $\rho$, stability threshold $\epsilon$, patience $T$, \#triples per neuron $K=16$.
    \STATE {\bfseries Output:} trained neuron weight $\vw$

    \STATE $\mu \gets 0$, $c \gets 0$
    \REPEAT
      \STATE $\vw \leftarrow \STEP(\vw)$
      \STATE $h \gets - \sum_{i=1}^{K} \tilde{\vw}_i \log \tilde{\vw}_i, \text{ where } \tilde{\vw}_i = \frac{\exp(\vw_i)}{\sum_j \exp(\vw_j)}$
      \STATE Increase $c$ by 1 if $|\mu - h| \le \epsilon$, else set $c \gets 0$
      \STATE $\mu \gets \rho \mu + (1-\rho) h$
      \IF{$c \ge T$}
        \STATE dominated $\gets$ check if $\exists i^*, \tilde{\vw}_{i^*} \ge 0.95$
        \STATE dispersed $\gets$ check if $\max_i \tilde{\vw}_i \le 0.4$
        \IF{dominated}
          \STATE resample all $(\vk_i, \vp_i, \vq_i)$ except the dominant one
          \STATE reset $\vw$ such that $\tilde{\vw}_{i^*}=0.9$ and $\tilde{\vw}_j=\frac{0.1}{K-1}$ for $j \neq i^*$
        \ELSIF{dispersed}
          \STATE resample all $(\vk_i, \vp_i, \vq_i)$
          \STATE reset $\vw$ such that $\tilde{\vw}_i=\frac{1}{K}$ for all $i$
        \ENDIF
      \ENDIF
      \STATE reset $c \gets 0$ if resampling is performed
    \UNTIL{training ends}
  \end{algorithmic}
\end{algorithm}

\begin{algorithm}[!h]
  \caption{Adaptive Discretization}
  \label{alg:discretization}
  \begin{algorithmic}
    \STATE {\bfseries Input:} weights $\mW$ returned by \cref{alg:resampling}, weight update function $\STEP$, EMA momentum $\rho$, stability threshold $\epsilon$, patience $T$.
    \STATE {\bfseries Output:} discretized Boolean network
    \REPEAT
      \STATE $\mW \gets \STEP(\mW)$
      \STATE $l \gets$ the first non-discretized layer
      \STATE Initialize $\mu_l \gets 0$, $c_l \gets 0$ if not initialized
      \STATE Compute average weight entropy $h_l$
      \STATE Increase $c_l$ by 1 if $|\mu_l - h_l| \le \epsilon$, else set $c_l \gets 0$
      \STATE $\mu_l \gets \rho \mu_l + (1-\rho) h_l$
      \IF{$c_l \ge T$}
        \STATE Freeze and discretize layer $l$ using \cref{eq:discretization}
        \STATE Mark layer $l$ as discretized
      \ENDIF
    \UNTIL{training ends}
  \end{algorithmic}
\end{algorithm}

\FloatBarrier

\section{Thermometer Encoding and Channel Visibility} \label{app:thermo_effects}

\subsection{Visual Effect on \cifar Channels} \label{app:thermo_visual}

\begin{figure*}[!h]
    \centering
    \includegraphics[width=0.72\linewidth]{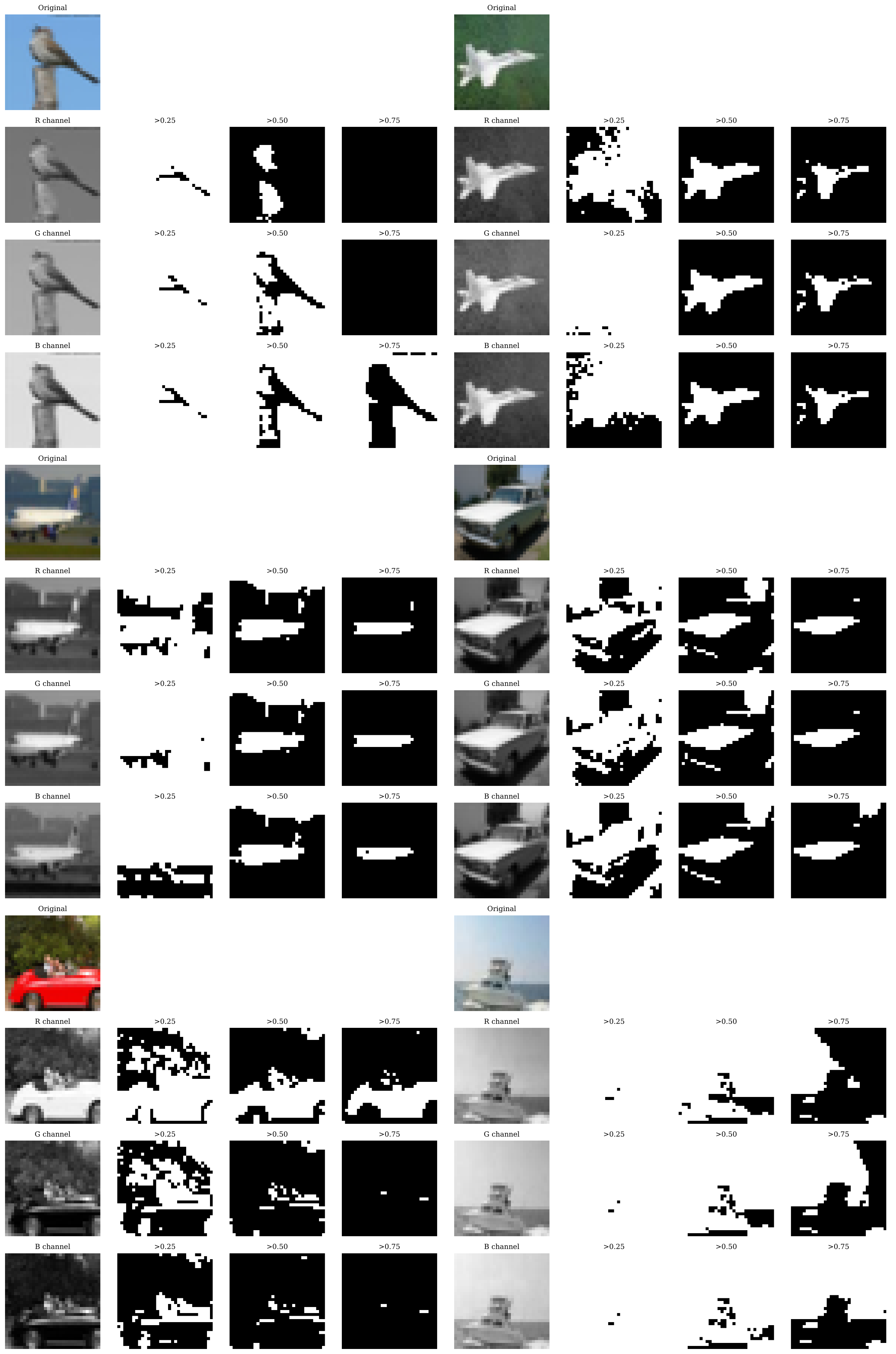}
    \caption{Illustration of the destructive effect of thermometer encoding on the channels of natural images from \cifar.}
    \label{fig:thermo_bird}
\end{figure*}

Thermometer encoding can substantially alter the channel structure of natural images. As shown in \cref{fig:thermo_bird}, many encoded channels lose color information and preserve only partial edge patterns. Moreover, only a small subset of channels contains non-trivial, non-overlapping signal. Often, one channel carries most label-relevant information, while other channels largely repeat it. This observation motivates the channel-visibility hypothesis in \cref{sec:exp_convolution}.

\subsection{\mnist Channel-Visibility Experiment} \label{app:mnist_channel_visibility}

\begin{figure}[!h]
    \centering
    \includegraphics[width=0.45\linewidth]{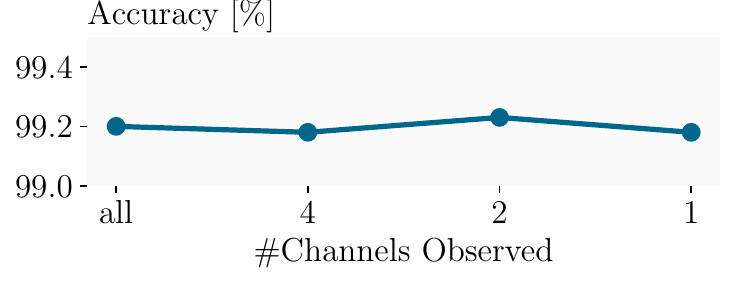}
    \caption{Channel-visibility experiment on \mnist. Restricting channel visibility leads to only negligible performance differences, unlike the clear improvement observed on \cifar.}
    \label{fig:mnist_channel_visibility}
\end{figure}

To test whether the channel-visibility paradox is tied to information loss from thermometer encoding, we repeat the channel-restriction experiment on \mnist. This setting is informative because thermometer encoding collapses \mnist inputs to a single Boolean channel, so it does not destroy cross-channel information. \Cref{fig:mnist_channel_visibility} shows that the paradox largely disappears: restricting channel visibility causes only negligible performance differences, rather than the substantial improvement observed on \cifar. This result supports the hypothesis that the channel-visibility paradox on natural images is driven by information loss across thermometer-encoded channels.

\FloatBarrier

\section{Variance Across Random Seeds} \label{app:variance_seeds}

To verify stability, we repeat the main connection-learning and convolution experiments from \cref{sec:exp_connection,sec:exp_convolution} across random seeds. \Cref{tb:DLGN_MNIST_rs,tb:DLGN_CIFAR_rs,tb:CLGN_MNIST_rs,tb:CLGN_CIFAR_rs} report the corresponding means and standard deviations. Our method consistently outperforms the baselines by a clear margin, indicating that the improvements are stable across seeds.

\input{tables/variance_seeds.tex}

\section{Additional Experiments and Ablations} \label{app:additional_ablation}

\subsection{\mnist Component Results}

We report the \mnist counterparts of the component studies from \cref{sec:exp_connection,sec:exp_convolution} in \cref{tb:DLGN_MNIST,tb:CLGN_MNIST}. The trends are consistent with the \cifar results in the main text: learned connections improve over DiffLogicNet, and the compact convolutional architecture improves over TreeLogicNet while using substantially fewer neurons.

\begin{table}[!h]
\centering
\caption{Comparison between DiffLogicNet and our method on \mnist.}
\label{tb:DLGN_MNIST}

\begin{tabular}{lccc}
\toprule
Method & Acc. & \emph{\#neurons} & \emph{\#BOPs} \\
\midrule
DiffLogicNet (small)    & $97.16\%$ & 48.0 K & 26.9 K \\  %
DiffLogicNet (medium)    & $98.80\%$ & 384 K & 214 K \\  %
\midrule
Ours (small)            & \textbf{$97.95\%$} & 48.0 K & 25.0 K \\  %
Ours (medium)            & \textbf{$99.10\%$} & 384 K & 190 K \\  %
\bottomrule
\end{tabular}
\end{table}

\begin{table}[!h]
\centering
\caption{Comparison between TreeLogicNet and our method on \mnist.}
\label{tb:CLGN_MNIST}

\begin{tabular}{lccc}
\toprule
Method & Acc. & \emph{\#neurons} & \emph{\#BOPs} \\
\midrule
TreeLogicNet-S	& $97.77\%$ & 185 K & 117 K \\
TreeLogicNet-M	& $98.96\%$ & 740 K & 427 K \\
TreeLogicNet-L	& $99.33\%$ & 11.8 M & 6.20 M \\
\midrule
Ours-S	    & \textbf{$99.15\%$} & 260 K & 128 K \\
Ours-M	    & \textbf{$99.35\%$} & 520 K & 246 K \\
\bottomrule
\end{tabular}
\end{table}

\subsection{Comparison with Fixed-Schedule Discretization Strategies} \label{app:fixed_schedule_discretization}

We compare the proposed adaptive discretization with two fixed-schedule discretization strategies and with a baseline that uses no training-phase discretization. All methods use the same training budget and are evaluated on two architectures.

\Cref{tab:discretization} shows that all training-phase discretization strategies improve over the no-discretization baseline by a clear margin. A well-tuned fixed schedule can approach the performance of adaptive discretization, but the adaptive strategy performs best in all cases without requiring an additional schedule search. The difference between the ``fixed every 10k'' and ``fixed every 20k'' columns shows that fixed-schedule discretization is sensitive to this predefined schedule.

\begin{table}[tbp]
\centering
\caption{Comparison between adaptive and fixed-schedule discretization strategies on Ours-T and Ours-S with seed variance. Reported results are test accuracy on \cifar.}
\label{tab:discretization}
\resizebox{0.9\columnwidth}{!}{
\begin{tabular}{lcccc}
\toprule
 & \begin{tabular}{c}Without\\ training-phase\\ discretization\end{tabular}
 & \begin{tabular}{c}Fixed Schedule\\ (every 10k)\end{tabular}
 & \begin{tabular}{c}Fixed Schedule\\ (every 20k)\end{tabular}
 & Adaptive \\
\midrule
Ours-T & 60.74\% $\pm$ 0.46\% & 62.20\% $\pm$ 0.15\% & 61.52\% $\pm$ 0.04\% & 62.92\% $\pm$ 0.31\% \\
Ours-S & 65.27\% $\pm$ 0.58\% & 66.57\% $\pm$ 0.21\% & 66.24\% $\pm$ 0.17\% & 66.87\% $\pm$ 0.30\% \\
\bottomrule
\end{tabular}
}
\end{table}

\subsection{Additional Results on \TIN and Other Modalities} \label{app:tinyimagenet_results}

We further evaluate our method on \TIN~\citep{le2015tinyimagenet,deng2009imagenet}. As shown in \cref{tab:tinyimagenet}, our large model achieves 23.54\% accuracy, outperforming TreeLogicNet-M (21.96\%) while using 4x fewer neurons. Our medium model achieves comparable accuracy to TreeLogicNet-M (20.78\% vs. 21.96\%) with 15x fewer neurons, and substantially outperforms TreeLogicNet-S (15.17\%) with 2x fewer neurons. These results extend the main vision experiments to a larger-scale dataset.

\begin{table}[tbp]
\centering
\caption{Test accuracy on \TIN. We compare our method with TreeLogicNet across model sizes.}
\label{tab:tinyimagenet}
\resizebox{0.4\columnwidth}{!}{
\begin{tabular}{lcc}
\toprule
Model & Acc. & \textit{\#Neurons} \\
\midrule
TreeLogicNet-S & 15.17\% & 3.28M \\
TreeLogicNet-M & 21.96\% & 26.2M \\
\midrule
Ours-M         & 20.78\% & 1.68M \\
Ours-L         & 23.54\% & 6.75M \\
\bottomrule
\end{tabular}
}
\end{table}

We also evaluate two non-image tasks: Breast Cancer classification for tabular data~\citep{wolberg1993breastuci,street1993nuclear} and ECG classification on the PhysioNet MIT-BIH Arrhythmia dataset for sequence data~\citep{goldberger2000physiobank,moody2001impact}. As shown in \cref{tab:other_modalities}, our method consistently outperforms DiffLogicNet under matched architectures, supporting the generality of the proposed connection-learning strategy beyond vision benchmarks.

\begin{table}[tbp]
\centering
\caption{Test accuracy on the Breast Cancer and PhysioNet MIT-BIH Arrhythmia (ECG classification) datasets. We compare our method and DiffLogicNet under identical model architectures.}
\label{tab:other_modalities}
\resizebox{0.65\columnwidth}{!}{
\begin{tabular}{llcc}
\toprule
Dataset & Method & Acc. & \textit{\#Neurons} \\
\midrule
\multirow{2}{*}{Breast Cancer} & DiffLogicNet & 73.33\% $\pm$ 0.13\% & 384 \\
 & Ours & 74.28\% $\pm$ 0.00\% & 384 \\
\midrule
\multirow{2}{*}{ECG} & DiffLogicNet & 97.53\% $\pm$ 0.03\% & 48K \\
 & Ours & 97.84\% $\pm$ 0.04\% & 48K \\
\bottomrule
\end{tabular}
}
\end{table}

\section{Implementation for TreeLogicNet} \label{app:convlogic}

Since \citet{petersen2024convolutional} did not release the official TreeLogicNet implementation, we use the community implementation from \citet{korinek_convlogic}. To validate this baseline, we compare its test accuracies with those reported by \citet{petersen2024convolutional}. \Cref{tb:convlogic} shows that the community implementation is slightly weaker under its default setting, but matches the reported TreeLogicNet-M accuracy after applying the same data augmentation used in our experiments. We therefore use this implementation as the TreeLogicNet baseline for convolutional Boolean networks.

\begin{table}
\centering
\caption{Comparison between the accuracy reported by \citet{petersen2024convolutional} and the community TreeLogicNet implementation, named ConvLogic.}
\label{tb:convlogic}
\resizebox{0.90\columnwidth}{!}{
\begin{tabular}{lccc}
\toprule
Model	& \citet{petersen2024convolutional} reported	& ConvLogic reported	&  ConvLogic + aug \\
\midrule
TreeLogicNet-S	& $60.38\%$	& $59.84\%$ 	& $59.79\%$ \\
TreeLogicNet-M	& $71.01\%$	& $69.15\%$		& $71.37\%$ \\
\bottomrule
\end{tabular}
}
\end{table}

\section{Bivariate Boolean Functions}\label{app:bool_fun}

\Cref{tb:all_two_input_boo_fun} lists the truth table for all sixteen bivariate Boolean functions. For example, $B_1$ is the constant-0 function, $B_2$ is AND, $B_4$ is identity, $B_8$ is OR, and $B_{16}$ is the constant-1 function.

\begin{table}[!h]
\centering
\caption{Truth table of the sixteen bivariate Boolean functions.} \label{tb:all_two_input_boo_fun}
\resizebox{\columnwidth}{!}{
\begin{tabular}{|c|c||*{16}{c|}}
\hline
$x_1$ & $x_2$  & $B_1$ & $B_2$ & $B_3$ & $B_4$ & $B_5$ & $B_6$ & $B_7$ &
$B_8$ & $B_9$ & $B_{10}$ & $B_{11}$ & $B_{12}$ & $B_{13}$ & $B_{14}$ & $B_{15}$ & $B_{16}$ \\
\hline\hline
0 & 0 & 0 & 0 & 0 & 0 & 0 & 0 & 0 & 0 & 1 & 1 & 1 & 1 & 1 & 1 & 1 & 1 \\
\hline
0 & 1 & 0 & 0 & 0 & 0 & 1 & 1 & 1 & 1 & 0 & 0 & 0 & 0 & 1 & 1 & 1 & 1 \\
\hline
1 & 0 & 0 & 0 & 1 & 1 & 0 & 0 & 1 & 1 & 0 & 0 & 1 & 1 & 0 & 0 & 1 & 1 \\
\hline
1 & 1 & 0 & 1 & 0 & 1 & 0 & 1 & 0 & 1 & 0 & 1 & 0 & 1 & 0 & 1 & 0 & 1 \\
\hline
\end{tabular}
}
\end{table}

\section{Pruning Algorithm} \label{app:pruning_algorithm}

We use the pruning algorithm in this appendix to compute the effective number of Boolean operations (\#BOPs) after training. This lightweight post-processing step removes trivial and disconnected neurons from the learned network, giving a more faithful estimate of Boolean-network inference complexity before hardware synthesis.

First, we exclude constant neurons (\ie $B_1$ and $B_{16}$), since they can share a single constant source in the circuit. We also remove identity neurons (\ie $B_4$ and $B_6$) by rewiring their inputs, and negation neurons (\ie $B_{11}$ and $B_{13}$) by composing the negation with adjacent operations. We then remove neurons that are not connected to the final output layer, since they do not affect the network output. \Cref{alg:pruning} gives the procedure. In our experiments, we use an equivalent GPU-based implementation to compute \#BOPs efficiently.

\begin{algorithm}[tb]
  \caption{Pruning}
  \label{alg:pruning}
  \begin{algorithmic}
    \STATE {\bfseries Input:} trained Boolean network
    \STATE {\bfseries Output:} pruned Boolean network, \#BOPs after pruning

    \STATE pruned $\gets$ output neurons $\cup \{0, 1\}$, BOPs $\gets$ 0
    \STATE Q $\gets$ queue(output neurons)
    \REPEAT
      \STATE current $\gets$ Q.dequeue()

      \IF{current is constant}
        \STATE connect to the constant neuron in the pruned network instead
      \ELSIF{current is identity or negation}
        \STATE $n \gets$ the nontrivial input neuron of current
        \STATE add $n$ to pruned and Q, respectively, if $n$ is not already in pruned
        \IF{current is negation}
          \STATE adjust the connection from $n$ to the neurons accepting current as their input to account for the negation
        \ENDIF
      \ELSE
        \STATE BOPs $\gets$ BOPs + 1
        \STATE $n_1, n_2 \gets$ two input neurons of current
        \STATE add $n_1, n_2$ to pruned and Q, respectively, if they are not already in pruned
      \ENDIF

    \UNTIL{Q is empty}
  \end{algorithmic}
\end{algorithm}

When comparing against DiffLogicNet and TreeLogicNet, we apply the same pruning procedure to all learned networks to ensure a fair comparison of \#BOPs. This pruning step is intentionally simpler than full logic synthesis, which can perform more advanced hardware-level simplification and mapping. Nevertheless, the resulting \#BOPs are comparable to those reported in \citet{petersen2022deep,petersen2024convolutional}, indicating that this procedure is sufficient for estimating Boolean-network complexity in the main comparisons.

\section{Experimental Details} \label{app:exp_details}

\paragraph{Setup} We evaluate our method on \mnist~\citep{lecun1998gradient} and \cifar~\citep{krizhevsky2009learning}. For \mnist, we round floating-point pixels to the nearest Boolean values and apply random rotations up to $15^\circ$ and random affine transformations with translations up to $10\%$ of the image size in each spatial direction during training. For \cifar, we use thermometer encoding with equal thresholds (see \cref{sec:background}) to convert floating-point pixels to Boolean strings of varying lengths, following \citet{petersen2022deep,petersen2024convolutional}. For convolutional networks, these Boolean strings are concatenated along the channel dimension. During training, we apply random horizontal flips and random $32 \times 32$ crops with 2-pixel padding. No data augmentation is used at inference. Both datasets use the population-count decoder described in \cref{sec:background}.

\paragraph{Baselines} For connection learning and convolution design, we compare against DiffLogicNet~\citep{petersen2022deep} and TreeLogicNet~\citep{petersen2024convolutional}, respectively. We use the official DiffLogicNet implementation and a community TreeLogicNet implementation that matches the reported results after applying our data augmentation protocol (see \cref{app:convlogic}), since the official TreeLogicNet implementation is not publicly available. For adaptive discretization, we compare against Gumbel noise injection~\citep{yousefi2025mind} using the official implementation. For fairness, all methods use the same input encoding, output decoding, data augmentation, and optimizer within each comparison.

\paragraph{Model Architectures}

We list the convolutional architectures used in the experiments below. The variants T (tiny), S (small), M (medium), and L (large) correspond to $k = 64, 128, 256,$ and $1024$, respectively. Let $N$ denote the number of thresholds in the thermometer encoding. For \mnist, we set $N=1$. For \cifar, we set $N=3$ for tiny and small models, $N=7$ for medium models, and $N=31$ for large models.

\begin{samepage}
\emph{\mnist Model Architecture:}
\begin{itemize}[leftmargin=*]
    \item \textbf{Layer 1:} Convolutional layer with kernel size $3\times3$, stride $2$, and padding $1$, mapping
    $(N, 28, 28) \rightarrow (k, 14, 14)$.
    
    \item \textbf{Layer 2:} Convolutional layer with kernel size $3\times3$, stride $1$, and padding $1$, mapping
    $(k, 14, 14) \rightarrow (k, 14, 14)$.
    
    \item \textbf{Layer 3:} Convolutional layer with kernel size $3\times3$, stride $2$, and padding $1$, mapping
    $(k, 14, 14) \rightarrow (4k, 7, 7)$.
    
    \item \textbf{Layer 4:} Convolutional layer with kernel size $3\times3$, stride $1$, and padding $1$, mapping
    $(4k, 7, 7) \rightarrow (4k, 7, 7)$.
    
    \item \textbf{Layer 5:} Flatten operation.
    
    \item \textbf{Layer 6:} Logic layer mapping $196k \rightarrow 625k$ neurons.
    
    \item \textbf{Layer 7:} Logic layer mapping $625k \rightarrow 625k$ neurons.

    \item \textbf{Layer 8:} Group-sum (population count) layer with $10$ output groups.
\end{itemize}
\end{samepage}

\begin{samepage}
\emph{\cifar Model Architecture:}
\begin{itemize}[leftmargin=*]
    \item \textbf{Layer 1:} Convolutional layer with kernel size $3\times3$, stride $2$, and padding $1$, mapping
    $(3N, 32, 32) \rightarrow (k, 16, 16)$.
    
    \item \textbf{Layer 2:} Convolutional layer with kernel size $3\times3$, stride $1$, and padding $1$, mapping
    $(k, 16, 16) \rightarrow (k, 16, 16)$.
    
    \item \textbf{Layer 3:} Convolutional layer with kernel size $3\times3$, stride $2$, and padding $1$, mapping
    $(k, 16, 16) \rightarrow (4k, 8, 8)$.
    
    \item \textbf{Layer 4:} Convolutional layer with kernel size $3\times3$, stride $1$, and padding $1$, mapping
    $(4k, 8, 8) \rightarrow (4k, 8, 8)$.

    \item \textbf{Layer 5:} Flatten operation.

    \item \textbf{Layer 6:} Logic layer mapping $256k \rightarrow 625k$ neurons.

    \item \textbf{Layer 7:} Logic layer mapping $625k \rightarrow 625k$ neurons.

    \item \textbf{Layer 8:} Group-sum (population count) layer with $10$ output groups.
\end{itemize}
\end{samepage}

\begin{table}[!h]
\centering
\caption{Comparison between Gaussian and residual initialization for regular models trained with our method.}
\label{tb:init_results}
\begin{tabular}{c c c c}
\toprule
Dataset & Model & Gaussian & Residual \\
\midrule
\multirow{2}{*}{\mnist}
 & small  & $97.95\%$ & $97.89\%$ \\
 & medium & $99.10\%$ & $98.61\%$ \\
\midrule
\multirow{3}{*}{\cifar}
 & small  & $56.13\%$ & $56.09\%$ \\
 & medium & $62.84\%$ & $63.52\%$ \\
 & large  & $64.53\%$ & $65.42\%$ \\
\bottomrule
\end{tabular}
\end{table}

\paragraph{Training Details}

We initialize regular models with Gaussian weights $\gN(0,1)$ to match DiffLogicNet. For convolutional models, we use the residual initialization from TreeLogicNet, setting the normalized weight of $B_4$ to $0.9$ and the normalized weights of the other Boolean functions to $0.1/15$. We also compare Gaussian and residual initialization for regular models on \mnist and \cifar. As shown in \cref{tb:init_results}, residual initialization does not consistently outperform Gaussian initialization for regular models, although it improves performance on the larger \cifar models.

All models are trained with Adam~\citep{kingma2015adam}. We use a constant learning rate of $0.01$ for regular models and $0.02$ for convolutional models, with batch size $128$. Learning-rate scheduling did not improve performance in our experiments, potentially because improvements in the relaxed model do not necessarily translate to improvements after discretization. Regular models are trained for 300K steps and evaluated every 1K steps to select the best checkpoint. Tiny, small, and medium convolutional models are trained for 600K steps and evaluated every 2K steps. The large convolutional model is trained for 700K steps with the same evaluation frequency. We do not use weight decay.

We initialize the connection-learning parameterization as follows. At initialization, $\vk = [1, \dots, 16]$, so each neuron starts with the 16 distinct Boolean operations listed in \cref{app:bool_fun}. For convolutional layers, we sample $(\vp, \vq)$ uniformly with replacement from the $3\times3$ receptive field. For non-convolutional layers, the candidate input set is much larger, so we initialize $(\vp, \vq)$ to encourage input coverage. Let $\din$ and $\dout$ denote the input and output dimensions of the layer. We assume $2 \times \dout \geq \din$. Otherwise, some input neurons cannot be considered by the final network. Let $r := \lfloor 2 \times \dout / \din \rfloor$. We assign all input neurons to the first $r \times \din$ input slots, with $r$ repetitions, sample the remaining slots uniformly with replacement from all input neurons, and then shuffle the slots. This process is repeated independently 16 times to initialize each entry of $(\vp, \vq)$. During resampling, $(\vp_i, \vq_i)$ are sampled uniformly from the receptive field for convolutional layers and from all input neurons for non-convolutional layers. We sample $\vk_i$ uniformly from $\{1, \dots, 16\}$.

Training times for all models are reported in \cref{tb:hyperparams}.

\paragraph{Training Hardware} All models are trained on a single RTX 2080 Ti GPU except the large convolutional model (Ours-L), which is trained on a single RTX 5090 GPU.

\paragraph{Hyperparameters}

We list detailed hyperparameter settings in \cref{tb:hyperparams}. For DiffLogicNet and TreeLogicNet, we use the hyperparameters from their original implementations.

For resampling, we use an exponential moving average (EMA) with momentum $\rho = 0.99$ and stability threshold $\epsilon = 5 \times 10^{-4}$ in all experiments. The resampling patience $T$ is tuned by model size, with larger models using larger patience values. Adaptive discretization is applied only to convolutional models, using the same EMA momentum $\rho = 0.99$ and stability threshold $\epsilon = 5 \times 10^{-4}$, with fixed patience $T = 200$.

For tiny, small, and medium convolutional models, resampling is enabled for the first $400\,\mathrm{K}$ training steps and is then disabled when adaptive discretization begins. For the large convolutional model, resampling is enabled for $500\,\mathrm{K}$ steps, followed by adaptive discretization for the remaining $200\,\mathrm{K}$ steps.

\begin{table}[h]
\centering
\caption{Hyperparameter settings and training time for regular and convolutional models.}
\label{tb:hyperparams}
\begin{tabular}{ccc c c c}
\toprule
Structure & Model & Dataset & Group Sum Temp. $\tau$ & Patience $T$ & Time (h)\\
\midrule
\multirow{5}{*}{Regular}
 & Ours (small)  & \mnist        & 10  & 100 & 2.1\\
 & Ours (medium) & \mnist        & 45  & 1000 & 8.3\\
 \cmidrule(l){2-6}
 & Ours (small)  & \cifar     & 33  & 100 & 3.7\\
 & Ours (medium) & \cifar     & 100 & 500 & 10.5\\
 & Ours (large)  & \cifar     & 100 & 1000 & 25.8\\
\cmidrule(lr){1-6}
\multirow{6}{*}{Convolutional}
 & Ours-S & \mnist     & 40  & 100 & 8.2\\
 & Ours-M & \mnist     & 63  & 100 & 16.0\\
 \cmidrule(l){2-6}
 & Ours-T & \cifar  & 20  & 100 & 5.7\\
 & Ours-S & \cifar  & 40  & 100 & 9.8\\
 & Ours-M & \cifar  & 63  & 100 & 18.6\\
 & Ours-L & \cifar  & 160 & 15000 & 23.2\\
\bottomrule
\end{tabular}
\end{table}

\section{Existing methods on reducing the inference cost}\label{app:current_effort_on_efficient_inference}

 We first stress that our model, namely Boolean networks, are distinct from binary neural networks (BNNs). 
Boolean networks are essentially Boolean circuits: 
the connections being sparse and weight-less, 
different nodes locally implementing different Boolean operations. 
Binary neural networks, in contrast, are neural networks with weighted connections ($-1$ and $1$). 
Each neuron locally performs the same operation: summing up incoming signals
by weighted sum, then outputs $-1$ or $1$ after passing through  a threshold function. 
Boolean networks realize different functions by way of changing the Boolean operation at each 
neuron, while BNNs realize different function by way of changing the weights of the edges.

The following two tables summarize the performance on \mnist and \cifar of existing methods 
of reducing the inference cost. 
\begin{table}[H]
\centering
\caption{Comparison of classification accuracy, Boolean-network complexity, and FPGA inference runtime on MNIST.}
\label{tab:mnist_fpga_runtime}

\begin{tabular}{lccc}
\toprule
Method & Acc. & \textit{\#BOPs} & FPGA runtime \\
\midrule
TTNet (small)~\citep{benamira2022scalable} & 97.23\% & 46 K & --- \\
TTNet (large)~\citep{benamira2022scalable} & 98.02\% & 360 K & --- \\
LUTNet~\citep{wang2020lutnet}              & 98.01\% & --- & 5 ns \\
DWN~\citep{bacellar2024differentiable}     & 98.77\% & --- & 45 ns \\
FINN CNV~\citep{umuroglu2017finn_table}    & 98.40\% & 5.28 M & 641 ns \\
FINN FCN~\citep{umuroglu2017finn_table}    & 98.86\% & 258 M & --- \\
LowBitNN~\citep{zhan2021field}             & 99.2\%  & --- & 152 $\mu$s \\
FPGA-NHAP~\citep{liu2022fpga}              & 97.81\% & --- & 4.9 ms \\
\midrule
TreeLogicNet-S & 97.77\% & 117 K & 6.96 ns \\
TreeLogicNet-M & 98.96\% & 427 K & 7.74 ns \\
TreeLogicNet-L & 99.33\% & 6.20 M & 8.76 ns \\
\midrule
Ours-S & 99.31\% & 130 K & 6.18 ns \\
Ours-M & 99.38\% & 251 K & 6.48 ns \\
\bottomrule
\end{tabular}

\end{table}

\begin{table}[H]
\centering
\caption{Comparison of classification accuracy, Boolean-network complexity, and FPGA inference runtime on CIFAR-10.}
\label{tab:cifar_fpga_runtime}

\begin{tabular}{lccc}
\toprule
Method & Acc. & \textit{\#BOPs} & FPGA runtime  \\
\midrule
Conv.\ TTNet (small)~\citep{benamira2022scalable} & 50.10\% & 0.57 M & --- \\
Conv.\ TTNet (large)~\citep{benamira2022scalable} & 70.75\% & 189 M & --- \\
LUTNet~\citep{wang2020lutnet}                     & 84.95\% & 1290 M & --- \\
XNOR-Net~\citep{rastegari2016xnor}                & 86.28\% & 1780 M & --- \\
FINN CNV~\citep{umuroglu2017finn_table}           & 80.10\% & 901 M &45.6 $\mu$s \\ %
RebNet (1 residual)~\citep{ghasemzadeh2018rebnet} & 80.59\% & 2270 M & 167 $\mu$s \\ %
RebNet (2 residuals)~\citep{ghasemzadeh2018rebnet} & 85.94\% & 2830 M & 333 $\mu$s  \\ %
\midrule
TreeLogicNet-S & 59.79\% & 0.52 M & 8.16 ns \\
TreeLogicNet-M & 71.37\% & 3.66 M & 9.06  ns\\
\midrule
Ours-T & 62.71\% & 0.08 M & 5.88 ns \\
Ours-S & 67.12\% & 0.16 M & 6.18 ns \\
Ours-M & 70.22\% & 0.32 M & 6.48 ns \\
Ours-L & 72.56\% & 1.08 M & 7.08  ns\\
\bottomrule
\end{tabular}

\end{table}

\section{Broader Impact}
\label{app:broader_impact}

This work aims to improve the inference efficiency of neural networks by learning compact Boolean networks that can be executed with simple Boolean operations. Potential positive impacts include lower inference latency, reduced memory and energy consumption, and improved deployability on resource-constrained hardware such as edge devices and FPGAs. These benefits may broaden access to efficient machine learning systems and reduce the environmental cost of large-scale inference.

The same efficiency gains can also lower the cost of deploying machine learning systems in settings where the downstream use may be harmful or insufficiently audited. In addition, because compact Boolean networks are still learned from data, they can inherit dataset biases and task-specific failure modes from the training distribution. We therefore view the contribution as a general efficiency technique rather than a substitute for application-specific safety, privacy, fairness, or robustness evaluation before deployment.

\section{LLM Disclosure}
\label{app:llm_disclosure}

This paper was assisted by PIRA~\citep{pira}, a research-assistant system powered by GPT-5.4/5.5 with high reasoning effort. The assistance included partial implementation assistance and writing polish. The authors are fully responsible for the presented final content.

%% file: tables/variance_seeds.tex
\begin{table}[!htbp]
\centering
\caption{Statistics on \mnist for connection learning.}
\label{tb:DLGN_MNIST_rs}
\begin{tabular}{lccc}
\toprule
Method & Acc. & \emph{\#neurons} & \emph{\#BOPs} \\
\midrule
DiffLogicNet (small)    & $97.03\% \pm 0.10\%$ & 48.0 K & 27.0 $\pm$ 0.1 K \\
DiffLogicNet (medium)   & $98.78\% \pm 0.02\%$ & 384 K & 215 $\pm$ 1 K \\
\midrule
Ours (small)            & \textbf{$97.77\% \pm 0.13\%$} & 48.0 K & 25.2 $\pm$ 0.4 K \\
Ours (medium)           & \textbf{$99.06\% \pm 0.03\%$} & 384 K & 190 $\pm$ 1 K \\
\bottomrule
\end{tabular}
\end{table}

\begin{table}[!htbp]
\centering
\caption{Statistics on \cifar for connection learning.}
\label{tb:DLGN_CIFAR_rs}
\begin{tabular}{lccc}
\toprule
Method & Acc. & \emph{\#neurons} & \emph{\#BOPs} \\
\midrule
DiffLogicNet (small)    & $52.70\% \pm 0.15\%$ & 48.0 K & 27.0 $\pm$ 0.7 K \\
DiffLogicNet (medium)   & $59.74\% \pm 0.11\%$ & 512 K & 311 $\pm$ 1 K \\
DiffLogicNet (large)    & $61.48\% \pm 0.76\%$ & 1.28 M & 780 $\pm$ 2 K \\
\midrule
Ours (small)            & \textbf{$55.81\% \pm 0.33\%$} & 48.0 K & 27.6 $\pm$ 0.3 K \\
Ours (medium)            & \textbf{$62.46\% \pm 0.27\%$} & 512 K & 296 $\pm$ 2 K \\
Ours (large)            & \textbf{$64.16\% \pm 0.26\%$} & 1.28 M & 655 $\pm$ 1 K \\
\bottomrule
\end{tabular}
\end{table}

\begin{table}[!htbp]
\centering
\caption{Statistics on \mnist for convolution.}
\label{tb:CLGN_MNIST_rs}
\begin{tabular}{lccc}
\toprule
Method & Acc. & \emph{\#neurons} & \emph{\#BOPs} \\
\midrule
TreeLogicNet-S    & $97.76\% \pm 0.04\%$ & 185 K & 120 $\pm$ 1 K \\
TreeLogicNet-M    & $98.87\% \pm 0.12\%$ & 740 K & 433 $\pm$ 2 K \\
TreeLogicNet-L    & $99.28\% \pm 0.03\%$ & 11.8 M & 6.20 $\pm$ 0.03 M \\
\midrule
Ours-S            & \textbf{$99.30\% \pm 0.02\%$} & 260 K & 131 $\pm$ 1 K \\
Ours-M            & \textbf{$99.35\% \pm 0.02\%$} & 520 K & 249 $\pm$ 1 K \\
\bottomrule
\end{tabular}
\end{table}

\begin{table}[!htbp]
\centering
\caption{Statistics on \cifar for convolution.}
\label{tb:CLGN_CIFAR_rs}
\begin{tabular}{lccc}
\toprule
Method & Acc. & \emph{\#neurons} & \emph{\#BOPs} \\
\midrule
TreeLogicNet-S    & $59.47\% \pm 0.35\%$ & 874 K & 538 $\pm$ 8 K \\
TreeLogicNet-M    & $71.30\% \pm 0.06\%$ & 7.00 M & 3.76 $\pm$ 0.06 M \\
\midrule
Ours-T	& \textbf{$62.92\% \pm 0.31\%$} & 145 K & 84.8 $\pm$ 0.7 K \\
Ours-S	& \textbf{$66.87\% \pm 0.30\%$} & 291 K & 164 $\pm$ 1 K \\
Ours-M	& \textbf{$69.98\% \pm 0.17\%$} & 582 K & 318 $\pm$ 1 K \\
Ours-L	& \textbf{$72.33\% \pm 0.23\%$} & 2.33 M & 1.08 $\pm$ 0.12 M \\
\bottomrule
\end{tabular}
\end{table}